\documentclass[sigconf,authorversion,screen,nonacm]{acmart}
\pdfoutput=1

\AtBeginDocument{%
  }

\setcopyright{acmlicensed}
\copyrightyear{2018}
\acmYear{2018}
\acmDOI{XXXXXXX.XXXXXXX}

\acmConference[Conference acronym 'XX]{Make sure to enter the correct
  conference title from your rights confirmation emai}{June 03--05,
  2018}{Woodstock, NY}
\acmISBN{978-1-4503-XXXX-X/18/06}

\usepackage{tabularx}
\usepackage{multirow}
\usepackage{dirtytalk}

\begin{document}

\title{Lying Blindly: Bypassing ChatGPT's Safeguards to Generate Hard-to-Detect Disinformation Claims}

\author{Freddy Heppell}
\affiliation{
    \institution{University of Sheffield}
    \department{Department of Computer Science}
    \city{Sheffield}
    \country{United Kingdom}
}
\email{frheppell1@sheffield.ac.uk}

\author{Mehmet E. Bakir}
\affiliation{
    \institution{University of Sheffield}
    \department{Department of Computer Science}
    \city{Sheffield}
    \country{United Kingdom}
}

\email{m.e.bakir@sheffield.ac.uk}
\author{Kalina Bontcheva}
\affiliation{
    \institution{University of Sheffield}
    \department{Department of Computer Science}
    \city{Sheffield}
    \country{United Kingdom}
}
\email{k.bontcheva@sheffield.ac.uk}

\renewcommand{\shortauthors}{Heppell et al.}

\begin{abstract}
As Large Language Models become more proficient, their misuse in coordinated disinformation campaigns is a growing concern. This study explores the capability of ChatGPT with GPT-3.5 to generate short-form disinformation claims about the war in Ukraine, both in general and on a specific event, which is beyond the GPT-3.5 knowledge cutoff. Unlike prior work, we do not provide the model with human-written disinformation narratives by including them in the prompt. Thus the generated short claims are \emph{hallucinations} based on prior world knowledge and inference from the minimal prompt. With a straightforward prompting technique, we are able to bypass model safeguards and generate numerous short claims. We compare those against human-authored false claims on the war in Ukraine from ClaimReview, specifically with respect to differences in their linguistic properties. We also evaluate whether AI authorship can be differentiated by human readers or state-of-the-art authorship detection tools. Thus, we demonstrate that ChatGPT can produce realistic, target-specific disinformation claims, even on a specific post-cutoff event, and that they cannot be reliably distinguished by humans or existing automated tools.

\end{abstract}

\maketitle

\section{Introduction}

Evolution in AI text generation technologies have significantly altered the landscape of disinformation dissemination and detection. Through their vast training datasets and human-in-the-loop reinforcement process, Large Language Models (LLMs) such as GPT-3.5 \citep{brown2020language} and GPT-4 \citep{OpenAI-GPT-4TechnicalReport-2023i} are able to generate fluent text in a given style and integrate world knowledge from training. This has well-known harmful applications in generating disinformation, already recognised with earlier, less-sophisticated models \citep{openaibettermodels, Buchanan-TruthLiesDisinformation-2021d}.

The ability of LLMs to output a mixture of reasoning based upon true information \citep{10.1145/3614321.3614325} and plausible falsehoods with equal confidence \citep{ren2023investigatingfactualknowledgeboundary}, as well as specific knowledge of conspiracies, hate groups \citep{McGuffie-RadicalizationRisksModels-2020p} and offensive stereotypes \citep{10.1145/3614321.3614325} equips them with great capacity to produce dangerous content \citep{mittelstadt2023protect}. To counter misuse, LLM providers try to implement safeguards \citep{OpenAI-GPT-4TechnicalReport-2023i}. 

Despite this, prior work has found that language models can be used to generate convincing long texts, by \emph{manipulating} human-written disinformation claims \citep{chen2023combating, Mosallanezhad-Topic-preservingSyntheticApproach-2020u, Stiff-DetectingComputer-generatedDisinformation-2022c, Vykopal-DisinformationCapabilitiesModels-2023c, lucas-etal-2023-fighting}. In these studies, models are being conditioned to generate disinformation by being provided with seed false narratives in the prompt. They are typically tasked to transform this seed narrative into an alternate format, e.g. from a human-authored disinformation claim to an entire article. Additionally, these claims often relate to well-established, pre-model-cutoff events. Researchers have also tried asking models explicitly to generate disinformation claims, but these are often refused due to safeguards \citep{Chen-LLM-generatedMisinformation-2023o,spitale23-disinforms-better}. Taken together, this leaves the open question whether LLMs are capable of creating disinformation narratives from scratch especially related to current events and without being provided with seed false narratives in the prompt.

To answer this open question, we investigate the ability of  ChatGPT with GPT-3.5 to generate disinformation on a given topic without seed false narrative prompting. We investigate both broad and narrow topics. For the former, the model is asked to generate short claims related to the war in Ukraine, while the latter narrow topic is specifically the Bucha massacre. The latter event happened after the ChatGPT3.5 cut-off date, so the model has no direct knowledge from its training data. Effectively, this exploits the model's hallucinations to generate disinformation.

We also demonstrate that GPT3.5's model safeguards can be bypassed with close to 100\% success rate, which allowed us to build a dataset of numerous AI-generated short disinformation claims. 

These AI-generated claims are compared against human-authored claims to demonstrate that ChatGPT is capable of generating convincing disinformation claims, which cannot be reliably distinguished from human-authored claims neither by human readers nor by AI authorship detection classifiers. One particularly important finding is that when constrained to a specific unknown event, the model successfully adapts its output to generate convincing claims. Our detailed analysis of the linguistic properties of both AI-generated and human-written claims did however uncover some general shortcomings of the AI outputs and differences induced by the change in subject.

In summary, this paper's contributions are: i) a dataset of human-written and ChatGPT-generated short false claims; ii) a detailed linguistic comparison between them; and iii) investigation of the ability of human readers and computational methods to identify which short claims are written by AI. The  summary statistics of the datasets are available publicly on GitHub\footnote{\anon[\url{https://anonymous.4open.science/r/chatgpt-ukraine-disinfo-F545/}]{\url{https://github.com/GateNLP/chatgpt-ukraine-disinfo}}}, and the complete disinformation datasets will be made available via Zenodo\footnote{\anon[For reviewing we have made them available anonymously at: \url{https://anonymous.4open.science/r/chatgpt-ukraine-disinfo-restricted-EF27/}]{\url{https://zenodo.org/records/10716080}}}.

\section{Related Work}

\subsection{Knowledge and Hallucinations of LLMs}

Language models are able to recall information present in training data and seemingly reason with it \citep{NEURIPS2022_9d560961}. However, any truthfulness is merely a product of it being a statistically likely output and just one of several factors optimised for in the reinforcement learning (from human feedback; RLHF) process \citep{mittelstadt2023protect}. The output of invented, yet context- and stylistically-appropriate, details is a type of `hallucination' \citep{zhang2023sirenssongaiocean}. Models can output factually correct and incorrect versions of the same information depending on the context \citep{zheng2023doeschatgptfallshort, jiang-etal-2024-large}, and have an extremely poor `perception' of factual boundaries so will readily output hallucinated falsehoods when exceeded \citep{ren2023investigatingfactualknowledgeboundary}. Recent events are particularly problematic, as they may postdate the knowledge cutoff, or predate it by an insufficient amount to be effectively used by the model \citep{cheng2024dateddatatracingknowledge}.

Furthermore, accurate recall of information from training data does not necessarily mean factually accurate outputs. As they are trained on vast quantities of web data, models may `accurately recall' content which is biased biased or untrue \citep{zhang2023sirenssongaiocean}, regarding hate groups and conspiracies \citep{McGuffie-RadicalizationRisksModels-2020p}, or contain offensive stereotypes \citep{10.1145/3614321.3614325}.

\subsection{Disinformation Risk of LLMs}

Significant concern about the possibility of disinformation generation was first raised with GPT-2, being used as a justification for the model's staged release \citep{openaibettermodels}. In practice state-of-the-art models had limited real-world impact due to factors such as the resources required to run them \citep{Buchanan-TruthLiesDisinformation-2021d}, and issues with quality and reliability, but these have been reduced the commercial availability of LLMs\citep{goldstein2023generativelanguagemodelsautomated}.

A significant risk factor of LLM disinformation is the ability to generate content at scale \citep{Buchanan-TruthLiesDisinformation-2021d, Kreps-NewsThatsMisinformation-2022l,goldstein2023generativelanguagemodelsautomated, Crothers-Machine-generatedTextMethods-2023n}, which could distort the legitimate popularity of viewpoints, or entirely undermine trust in general by overwhelming people with poor quality information \citep{chen2023combating}. The latter is similar to the `firehose of falsehood' strategy used in contemporary Russian propaganda; the large-scale dissemination poor-quality, potentially conflicting information aiming to confuse and overwhelm an audience \citep{Paul-RussianFirehoseCounter-2016g}. LLMs also permit new forms of disinformation, for example re-targeting a claim to appeal to multiple distinct audiences or even specific individuals \citep{goldstein2023generativelanguagemodelsautomated, augenstein2023factualitychallengeseralarge}, and reducing the risk of using unusual phrases indicating an unfamiliarity with the language or culture.

Multiple cases of widespread LLM-generated content have already been uncovered, including thousands of autonomous false news sites\footnote{\url{https://www.newsguardtech.com/special-reports/ai-tracking-center/}} and large networks of ChatGPT-powered conversational bots on social media\footnote{\url{https://ab.co/4dfoOgB}}.

LLM-generated disinformation appears to be similarly effective to human-authored disinformation. In different studies it has been found to be perceived as more convincing \citep{spitale23-disinforms-better}, or perceived as less accurate but equally as likely to be shared \citep{Bashardoust-ComparingWillingnessNews-2024f}.

\subsection{Disinformation Generation with Language Models}

Much of the work studying disinformation generation has focused on the generation of `end-product' disinformation, i.e. a disinformation article, and provided the disinformation itself. The model may be provided with a specific subject matter \citep{Mosallanezhad-Topic-preservingSyntheticApproach-2020u, Huang-HarnessingPowerExplanation-2023u}, article headlines \citep{uchendu-etal-2020-authorship}, headlines accompanied by metadata \citep{Zellers-DefendingAgainstNews-2019c, Stiff-DetectingComputer-generatedDisinformation-2022c} or detailed abstracts \citet{Vykopal-DisinformationCapabilitiesModels-2023c}, or true information to perturb \citep{lucas-etal-2023-fighting}. These are summarised in table \ref{tab:prompt_examples}, with examples of the level of information specified.

Some studies have attempted generation without seeding with substantial narratives, but were thwarted by safeguards \citep{Chen-LLM-generatedMisinformation-2023o}. \citet{spitale23-disinforms-better} used simple descriptions of a claim but all related to well-established conspiracies. Generation is often in the style of longer-form news content \citep{Mosallanezhad-Topic-preservingSyntheticApproach-2020u, Zellers-DefendingAgainstNews-2019c, Vykopal-DisinformationCapabilitiesModels-2023c, Chen-LLM-generatedMisinformation-2023o, uchendu-etal-2020-authorship, Huang-HarnessingPowerExplanation-2023u}, short-form social media posts \citep{spitale23-disinforms-better}, or both \citep{Stiff-DetectingComputer-generatedDisinformation-2022c}. One study found that the outputs can include convincing novel (i.e. not supplied in the prompt) arguments in favour of the disinformation \citep{Vykopal-DisinformationCapabilitiesModels-2023c}.

To our knowledge, no prior work has directly assessed generated disinformation claims outside of long-form content, and no work has generated disinformation successfully without specifying the narrative.

LLMs should be prevented from outputting harmful content by safeguards, which can be intrinsic (through RLHF) or extrinsic (filtering inputs and outputs of the model) methods. While these methods may stop a cursory attempt to produce offensive content, it is possible to bypass them using approaches with a range of sophistication \citep{wei23-jailbroken}. A common strategy to influence model behaviour is to give it a role or persona, such as a news curator or journalist for disinformation tasks \citep{lucas-etal-2023-fighting}. This has also been shown to improve performance where the persona is an appropriate expert \citep{salewski23-impersonation}, although in some cases safeguards may be increased or affected in unpredictable ways \citep{li2024chatgptdoesnttrustchargers}. Safeguards can also be influenced by justifications as to why a behaviour is appropriate (e.g for academic purposes) \citep[\S6]{OpenAI-GPT-4TechnicalReport-2023i}.

\begin{table*}[]
    \centering
    \caption{Examples of information provided in prior work about disinformation generation.}
    \label{tab:prompt_examples}
    \begin{tabularx}{\textwidth}{@{}lX@{}}
        \toprule
        Study & Example of information given in prompt \\
        \midrule
        \citet{Mosallanezhad-Topic-preservingSyntheticApproach-2020u} & ``the wedding of prince harry and meghan markle''  \\
        \citet{Huang-HarnessingPowerExplanation-2023u} & ``New York was hit by a large-scale terrorist attack'' \\
        \citet{uchendu-etal-2020-authorship} & ``Putin and Xi are using the coronavirus crisis to extend their control. Across the world, Trump is struggling to keep up'' \\
        \citet{Zellers-DefendingAgainstNews-2019c} & ``New Research Shows that  Vaccines Cause Autism''$^\#$ \\
        \citet{Stiff-DetectingComputer-generatedDisinformation-2022c} & Following \citet{Zellers-DefendingAgainstNews-2019c} $^\star$ 
        \\
        \citet{Vykopal-DisinformationCapabilitiesModels-2023c} & ``People die after being vaccinated against COVID-19''. \\
        \citet{lucas-etal-2023-fighting} & ``San Francisco had twice as many drug overdose deaths as COVID deaths last year. This true state of emergency is met with political indifference if not encouragement.'' \\
        \citet{spitale23-disinforms-better} & ``climate change is real'' \\
        \citet{Chen-LLM-generatedMisinformation-2023o} & ``healthcare'' \& ``fake news'' $^\star$ $^\dagger$ \\
        \midrule
        Ours & ``Russia Ukraine war disinformation'', \newline ``Russia Ukraine war disinformation about the Bucha Massacre and different banned weapons use'' \\
        \bottomrule
    \end{tabularx}
    {\footnotesize
         For other works, if multiple levels of specificity were used, the least specific is chosen. $^\#$metadata was also provided. $^\star$also used unconstrained generation.
         $^\dagger$high rate (>90\%) refusal.
    }
\end{table*}

\subsection{Detection by Machines}

Numerous approaches have been proposed to detect AI-generated text (see surveys \citep{Crothers-Machine-generatedTextMethods-2023n, uchendu23_attributionreview} generally and \citep{chen2023combating} specifically for disinformation). Broadly, these approaches consider statistical methods (i.e. the likelihood of a model outputting the given text), supervised classification with transformers, or hybrids of these approaches, or inserting imperceptible watermarks into output \citep{pmlr-v202-kirchenbauer23a}. Commercial detection products, such as ZeroGPT\footnote{\url{https://www.zerogpt.com}}, claim high accuracy but do not publicise their methods. 

In practice, these methods are generally ineffective \citep{sadasivan2024aigeneratedtextreliablydetected}. Performance is dependent on length\footnote{Most work uses longer-form content, or at least several sentences \citep{uchendu23_attributionreview}} \citep{ippolito-etal-2020-automatic}, the exact training data used \citep{Tourille-AutomaticDetectionTweets-2022q}, and domain similarity \citep{Bakhtin-RealFakeText-2019r}. Numerous effective attack methods have been proposed \citep{wu2024surveyllmgeneratedtextdetection, macko2024authorshipobfuscation}, including as simple as inserting a single space character \citep{cai2023evadechatgptdetectorssingle}. Their use in education has been criticised due to poor performance \citep{Weber-Wulff-TestingDetectionText-2023a} and potential for bias against non-native speakers \citep{2023nonnativewriters}. One commercial tool, created by OpenAI themselves, was discontinued after 6 months due to ``low rate of accuracy''\footnote{\url{https://openai.com/index/new-ai-classifier-for-indicating-ai-written-text/}}.

\subsection{Detection by Humans}

Humans struggle to distinguish LLM and human-authored texts in general \citep{brown2020language, Clark-Human-Evaluation-2021f, Kreps-NewsThatsMisinformation-2022l} and specifically in relation to disinformation \citep{spitale23-disinforms-better, Chen-LLM-generatedMisinformation-2023o, Kreps-NewsThatsMisinformation-2022l}.

Some improvement is gained from training annotators \citep{ippolito-etal-2020-automatic, Clark-Human-Evaluation-2021f} and some individuals appear to be highly skilled at this task \citep{ippolito-etal-2020-automatic}. Whereas machine detectors exploit statistical patterns present in long, fluent text, human annotators identify structural or factual errors \citep{ippolito-etal-2020-automatic, Clark-Human-Evaluation-2021f}. Longer texts allow more room for errors such as topic drift, it may be easier to identify implausible or incoherent facts when they are the sole content.

\section{Generating the Claim Dataset}

In this section, we will discuss obtaining human-authored baseline data (\S \ref{sec:human_data}), using ChatGPT to generate similar data (\S\ref{sec:prompt}-\ref{sec:gpt_data}), and how this data fits within the established definitions of disinformation (\S\ref{sec:truthfulness}).

\begin{figure*}
    \centering
    \scalebox{0.9}{
    \fbox{
    \begin{tabularx}{1.05\textwidth}{@{}lX@{}}
        ClaimReview & ``Ukraine has been shooting residents of Donetsk and Lugansk just because they wanted to speak Russian.'' \\
        ChatGPT-General & The Ukrainian government is deliberately provoking the conflict in order to secure military and financial aid from the West.  \\
        ChatGPT-Topic & The Bucha Massacre was a result of Ukrainian forces acting on misinformation provided by Ukrainian politicians and military leaders.\\ 
    \end{tabularx}
    }
    }
    \caption{Examples of claims from each dataset}
    \label{fig:dataset_ex}
\end{figure*}

\subsection{Human-Authored Claims}\label{sec:human_data}
The Fact Check Data Feed\footnote{\url{https://www.datacommons.org/factcheck/download}}, commonly referred to as \emph{ClaimReview} after the data schema it follows, includes over 50,000 fact-checking articles from more than 1,000 fact-checking organisations in multiple languages. The feed is derived from data submitted by fact-check authors, so avoids data accuracy issues that may result from automated scraping, but does not include some notable fact-checkers such as Snopes or EUvsDisinfo. Each debunk article includes a concise summary of the evaluated claim and additional metadata.

We first identified 23,040 English-language claims using the fastText language identification model \citep{joulin2016fasttextzip, joulin-etal-2017-bag}, then used BERTopic \citep{grootendorst2022bertopic} to perform unsupervised topic clustering. We used a SentenceTransformers MiniLM model \texttt{all-MiniLM-L6-v2} \citep{reimers-gurevych-2019-sentence, wang-minilm-2020}, with a minimum topic size of 10, and an automatically determined, unbounded number of clusters with no.

283 clusters were formed containing 13,055 claims; 9,985 claims did not have enough close neighbours to form clusters meeting the minimum size. After manual examination, 4 were deemed relevant to the war in Ukraine, containing 312 claims, of which 282 were retained after a manual check. These claims form our \emph{ClaimReview} baseline of human-authored fact-check claims.

Within this subset, 49\% of claims originate from PolitiFact. This is likely because they are the largest author of English-language claims (23\%), and appear to write more frequently about the war; a simple keyword search of conflict-related terms returns 41\% PolitiFact authorship. The remaining 51\% of claims are from 16 sources including some non-anglosphere organisations. As all claims are short statements of `fact', we do not believe there is sufficient linguistic scope for organisational style to affect the claims\footnote{There is no significant correlation between LIWC attributes and whether a claim is from PolitiFact, following methodology in section \ref{sec:liwc_method}.}. 

Assessing the truthfulness of these claims, even given the accompanying fact-check, is complex, as there is no standardised rating system within the ClaimReview markup. However it is known that fact-checkers tend to review `dubious' claims which are more likely to be false, resulting in an unequal distribution of output labels \citep{Lee2023-vp}\footnote{\citep{Lee2023-vp} found this holds for PolitiFact, and that Snopes, which is not included in Factcheck Datafeed, had the highest proportion of true claims.}. Therefore it is likely that the majority of included claims are either rated false or appear to be false until properly researched, and very few are obviously true.

\subsection{Prompt Design and Evaluation}\label{sec:prompt}
All ChatGPT texts were generated with the \texttt{gpt-3.5-turbo} model.

Firstly, three prompting strategies were evaluated using both the UI and API: asking ChatGPT to produce (i) a single or (ii) multiple instances of disinformation, and (iii) asking for multiple instances with a safeguard-bypass strategy. Each prompt was trialled 5 times in isolated sessions.

To bypass safeguards, we modify the system prompt to inform the model it is an assistant whose purpose is to help understand disinformation content, and the prompt asks the model to describe the actions of a third party (a fact-checker) instead of directly producing the content. 

As in prior work \citep{Chen-LLM-generatedMisinformation-2023o, spitale23-disinforms-better}, ChatGPT always refused to generate disinformation when asked plainly through the UI (although the API had a highly inconsistent rejection rate), but our bypass strategy was always successful in both the UI and API. All our prompts contained the word `disinformation', corroborating prior suggestions that safeguards are not always activated by words clearly related to malicious content   \citep{Chen-LLM-generatedMisinformation-2023o}.

The data was first obtained in April 2023, although a recent trial showed they still function, through the web UI or API via the official OpenAI Python library.

As the prompts only require specifying a vague topic (opposed to other works, where prompts require human-authored input data), there is significant potential for misuse. The prompts, unless used for research, violate OpenAI's usage policy\footnote{\url{https://openai.com/policies/usage-policies}}. As such, they will only be shared on request with academic researchers. 

\subsection{Dataset Creation}\label{sec:gpt_data}

The final dataset was generated with the ChatGPT API (see examples in figure \ref{fig:dataset_ex}). The prompt with safeguard bypass was used, along with an initial system prompt, to generate over 300 disinformation claims (henceforth \textbf{ChatGPT-General}) by repeatedly sending ``More examples...'' in the same conversation. Additionally, the prompt was modified to ask specifically for claims regarding the Bucha massacre, an event after its knowledge cutoff, and use of banned weapons in war (\textbf{ChatGPT-Topic}). 

At the time of generation, the knowledge cutoff date of GPT-3.5's training corpus was September 2021, so it had no prior exposure to the current war in Ukraine, which begun in February 2022.

In the latter prompt, no information about the Bucha massacre was provided beyond the name, so the claims generated for this prompt are the product of the model's reasoning ability, not direct knowledge.

Responses were processed to remove any introductory text, numbering of the claims, and exact duplicate claims. 282 claims were randomly sampled from each, to match the size of the ClaimReview subset. No manual screening for quality or fluency was performed.

\subsection{Truthfulness of Claims}\label{sec:truthfulness}
We do not assess the truthfulness of these claims for several reasons.

Claims generated by ChatGPT may be based upon world knowledge from its training data, but since they are describing a post-cutoff event,  they clearly cannot integrate any direct knowledge. In fact, the claims generated would be the same as if the 2022 invasion of Ukraine had not happened.

We argue classifying this as disinformation is consistent with established definitions (see review by \citet{Fallis-WhatDisinformation-2015p}), which recognise the possibility of disinformation to be accurate yet intentionally misleading. Furthermore, a risk-factor of LLM disinformation is its ability to scale \citep{Buchanan-TruthLiesDisinformation-2021d, Kreps-NewsThatsMisinformation-2022l}, consistent with the established propaganda strategy \emph{firehosing}: the large-scale dissemination poor-quality, potentially conflicting information aiming to confuse and overwhelm an audience \citep{Paul-RussianFirehoseCounter-2016g}. \citeauthor{Fallis-WhatDisinformation-2015p} observes that, while it does not fall within his proposed definition, it may be desirable to include information that ``is not misleading [...], but is likely to keep people in ignorance'' \citep{Fallis-WhatDisinformation-2015p}, which we suggest would include firehosing.

Furthermore, as the generated claims tend to lack specific details (see section \ref{sec:liwc_exps}), they may not have enough grounding in specific events or contain enough details to be verified.

\section{Linguistic Comparison}\label{sec:liwc_exps}
In order to determine whether ChatGPT can generate disinformation similar to human-written false claims, we compare the three datasets using the LIWC-22 \citep{liwc22}. LIWC calculates the overlap between words in a target text and over a hundred\footnote{See \citep{liwc22dev} for a complete inventory} dictionaries of words and word stems, representing linguistic dimensions and social and psychological constructs, and has been shown to provide consistent and valid representations of text characteristics \citep{liwc22dev}.

The majority of the metrics represent the percentage of target words present in the given dictionary or its child dictionaries, e.g. a text where 5\% of words are related to anger and 5\% sadness would be given \emph{Anger}=5, \emph{Sadness}=5, \emph{Emotion}=10. Additional summary variables provide descriptive statistics of the text, e.g. words per sentence, emotional tone.

Two comparisons are made: i) between human and ChatGPT-authored claims to investigate potential stylistic differences between authoed texts; ii) between ChatGPT-General and ChatGPT-Topic claims to investigate the effect that topic-specificity had.

\subsection{Analysis Methodology}\label{sec:liwc_method}

LIWC metrics were calculated each claim text independently. For the first comparison, each claim is assigned a boolean label indicating if it originated from ClaimReview or either of the two ChatGPT datasets. For the second comparison, ChatGPT-authored texts were assigned a boolean label indicating if they originated from ChatGPT-General or ChatGPT-Topic, with ClaimReview texts excluded.

The correlation between each LIWC metric and the boolean labels is calculated with Point Biserial Correlation, a special case of Pearson correlation when one variable is boolean \citep{pbcorr}. For each comparison, the correlation $r$ and two-sided $p$-value was calculated between each LIWC metric $m_i$ and the boolean label $y$. Metrics with constant value (such as Emoji, which was always 0) were excluded as the correlation is undefined in this case. We present all correlations as positive by calculating both inverse assignments of $y$; when $y$ is mutually-exclusive the correlations are symmetrical, e.g. if the correlation between $m_i$ and human authorship is $-r$, the correlation between $m_i$ and ChatGPT authorship will be $+r$. Only correlations $r \geq 0.3$ are included, all of which have $p \ll 0.01$.

\subsection{Comparison of Claim Authorship}
\begin{table}[]
    \centering
    \caption{Correlation of LIWC metrics to human and ChatGPT authorship}
    \scalebox{0.9}{
    \begin{tabular}{l r l r}
        \toprule
        \multicolumn{2}{c}{Human} & \multicolumn{2}{c}{ChatGPT} \\ \cmidrule(r){1-2}\cmidrule(l){3-4}
        Metric & $r$ & Metric & $r$  \\
        \midrule
        Other Punctuation & 0.597 & Causation & 0.512 \\
        All Punctuation & 0.591 & Words per Sentence & 0.418 \\
        Communication & 0.352 & Dictionary Words & 0.418 \\
        Numbers & 0.350 & Cognitive processes & 0.391 \\
        Apostrophes & 0.347 & Space & 0.368 \\
        Personal Pronouns & 0.344 & Cognition & 0.334 \\
        Time & 0.323 & Articles & 0.333 \\
        Social processes & 0.317 && \\
        Male references & 0.312 && \\
        Adverbs & 0.306 && \\
        \bottomrule
    \end{tabular}
    }
    \label{tab:liwc_human_vs_chatgpt}
\end{table}

Correlations of LIWC attributes to human and ChatGPT authorship are shown in table \ref{tab:liwc_human_vs_chatgpt}.

Human-authored texts appear to include more specific details, such as numbers and terms related to time. They also appear to use quotations more frequently, which presents as a correlation to other punctuation, caused by quote marks, communication words (e.g. ``said''), and is also likely the explanation for a correlation between ChatGPT texts and higher words-per-sentence but not word count, meaning the texts are of similar length but if human-authored they are broken into more sentences. ChatGPT texts tend to use articles (e.g. ``a'', ``an'') more frequently, which appears to be from both the common usage of phrases such as ``a result of'', and human texts following the journalistic style in which articles tend to be omitted (e.g. ``video shows'' over ``a video shows'').

ChatGPT's texts do not ever make reference to individuals, even in abstract terms (e.g. ``president''), instead referring to nations or non-specific groups (e.g. ``separatists''), causing the correlation of personal pronouns to human texts. Additionally this is supported by the correlation between ChatGPT texts and dictionary words (words in any metric dictionary), as proper nouns are unlikely to feature in them. ChatGPT's texts appear to be much more formal, lacking contractions as indicated by \emph{Apostrophes}, and are more explanatory (indicated by \emph{Causation}, as they are often structured as ``\emph{X} did \emph{Y}, causing \emph{Z}'')

\begin{table}[]
    \centering
    \caption{Correlation of LIWC metrics of ChatGPT-authored texts to topic specificity}
    \scalebox{0.9}{
    \begin{tabular}{l r l r}
        \toprule
        \multicolumn{2}{c}{ChatGPT-General} & \multicolumn{2}{c}{ChatGPT-Topic} \\ \cmidrule(r){1-2}\cmidrule(l){3-4}
        Metric & $r$ & Metric & $r$  \\
        \midrule
        Present focus & 0.545 & Causation & 0.599 \\
        Politics & 0.414 & Cognition & 0.552 \\
        Total function words & 0.366 & Cognitive processes & 0.548 \\
        Linguistic Dimensions & 0.360 & Past focus & 0.500 \\
        Common adjectives & 0.358 & Death & 0.398 \\
        Interpersonal conflict & 0.314 & Physical & 0.386 \\
        Social behaviour & 0.303 & Power & 0.311 \\
        \bottomrule
    \end{tabular}
    }
    \label{tab:liwc_general_topic}
\end{table}

\subsection{Comparison of Topic Specificity}
Correlations of LIWC attributes to whether the text was generated with the general or topic-specific prompt are shown in table \ref{tab:liwc_general_topic}. There are some slight stylistic differences between the texts, but no clear structural differences.

Asking ChatGPT to focus on a specific event caused it to switch from present to past tense (there is a correlation between ChatGPT-General and \texttt{focuspresent}, and ChatGPT-Topic and \texttt{focuspast}), matching the more frequent tense in human texts.

There is a clear shift in subject matter with the general claims discussing political factors and broad conflict-related terms (\emph{Interpersonal conflict}, e.g. \emph{``fight''}, \emph{``attack''}; its parent \emph{Social Behaviour}), to the topic-specific claims more concretely discussing consequences (\emph{Death}, e.g. \emph{``die''}, \emph{``kill''}) and rationale (\emph{Cognition}, specifically its child \emph{Causation}; \emph{Power}, e.g.\emph{``order''}, \emph{``allow''}).

The correlation to the linguistic variables (\emph{Common adjectives}; its parent \emph{Linguistic Dimensions} which includes all linguistic function words) appears to be an artifact of the LIWC dictionaries; no such correlation exists compared to the proportions of adjectives calculated by a statistical POS tagger.

\subsection{Keyword Similarity}

\begin{table}[]
\caption{Top 1\% most frequent words per-class with c-TF-IDF weighting}
    \begin{center}
\begin{tabular}{lll}
\toprule
ClaimReview & ChatGPT-General & ChatGPT-Topic \\
\midrule
\bfseries president & government & forces \\
ukraine & \itshape eastern & used \\
\bfseries video & fighting & \itshape result \\
\bfseries trump & conflict & \bfseries massacre \\
\bfseries biden & crimea & bucha \\
\bfseries shows & \itshape annexation & \itshape backed \\
russia & right & weapons \\
\bfseries putin & \itshape separatists & banned \\
\bfseries joe & using & \itshape separatist \\
\bfseries company & \bfseries establish & ukrainian \\
\bfseries said & \itshape necessary & \itshape use \\
aid & ukrainian & russian \\
\bfseries photo & \itshape protect & way \\
\bfseries says & \itshape region & \itshape cluster \\
\bfseries money & ukraine & bombs \\
did & \bfseries planning & \itshape phosphorus \\
 & \bfseries threat & white \\
 & state & military \\
 & \itshape interests & \itshape civilian \\
 & \itshape harm & \bfseries sides \\
\bottomrule
\end{tabular}
\end{center}
\footnotesize\textbf{bold}: word is exclusive to this dataset.\\\textit{italic}: word is only used in ChatGPT-authored datasets.

    \label{tab:ctfidf_freqs}
\end{table}

To evaluate the differences in topics and entities discussed in each dataset, we calculate the c-TF-IDF-weighted term frequency in each dataset. This metric is an adaptation of TF-IDF to consider the significance of terms within a subset of a corpus determined by class labels, instead of a single document \citep{grootendorst2022bertopic}. Additionally, we use the square-root of the normalised term frequency to reduce the weight impact of highly frequent terms\footnote{This is included in the implementation of \citep{grootendorst2022bertopic}} \citep{tfsqrt}. Unweighted term frequencies were calculated with the Spacy English tokenizer with stopwords excluded. The weighted frequency $W_{t,c}$ of term $t$ in class $c$ is defined in equation \ref{eq:ctfidf}, where $\operatorname{tf}_{t,c}$ is the frequency of term $t$ in class $c$; $\|X\|_1$ is the L1 norm of $X$; $A$ is the mean number of terms per class; and $\operatorname{tf}_t$ is the frequency of term $t$ across all classes. 

\begin{equation}\label{eq:ctfidf}
    W_{t,c} = \sqrt{\|\operatorname{tf}_{t,c}\|_1} \cdot \log\left(1+\frac{A}{\operatorname{tf}_t}\right)
\end{equation}

The highest 1\% of c-TF-IDF-weighted frequencies\footnote{A proportion is shown due to the differing number of unique terms per dataset.} for each dataset, and whether these terms are exclusive to the given dataset or ChatGPT-authored datasets, are shown in table \ref{tab:ctfidf_freqs}.

ChatGPT-General is characterised by words related to the 2014 unrest in Ukraine including the annexation of Crimea and the formation of separatist forces in the east of the country (``Crimea'', ``annexation'', ``eastern'', ``separatists''), as the current conflict began after the model's knowledge cutoff.

However, in ChatGPT-Topic, words which reflect the focus of the prompt are highly frequent  (``bucha'', ``massacre'', ``cluster'', ``bombs'',  ``white'', ``phosphorus''). Although many of these terms are not exclusively used in ChatGPT-Topic, their high weighted frequency indicates that they are used significantly more often (e.g. ``bucha'' appears in ChatGPT-Topic 92 times and ClaimReview once). ``Crimea`` is never used and other 2014-related keywords are less frequent, suggesting the model correctly substitutes explicit references to the 2014 unrest with the Bucha massacre event, despite this being unknown to it. The usage of ``sides'' is exclusively from the phrase ``both sides'' in approximately 10\% of claims in an apparent effort to make more balanced allegations, but this is not seen in ChatGPT-General.

World knowledge not included in the prompt (e.g. types of banned weapons) is still used, but without explicit reference to historical events. This has concerning implications for LLM-generated disinformation, as it suggests the knowledge cutoff does not limit disinformation capability.

None of the generated texts contain references to individuals directly or indirectly (e.g. ``trump''/``biden''/``putin'' or ``president``), including when discussing the annexation of Crimea, for which the model is aware of the respective leaders of the involved nations. This is in contrast to ClaimReview data, which frequently relate to specific individuals and their offices. This may be due to the model's safety alignment. Furthermore, the human-authored claims frequently contain references to photos and videos, but none of the ChatGPT-authored claims do.

ChatGPT-generated claims appear to be less likely to discuss specific events, as seen by the significance of terms such as ``planning'' (which is unlikely to appear in a real fact-check as the claim may be unverifiable, and never appears in our ClaimReview subset), and ``necessary'' which is used in the context of alleging motivation (``The annexation of Crimea was necessary to...''). This is in contrast to the ClaimReview set using terms such as ``says'' to allege a specific statement by an individual, or ``shows'' referring to a specific ``photo'' or ``video''.

The ChatGPT-generated texts are of a similar average length (general: 21 tokens, topic: 19) to the human-written claims (19).

\section{Human AI Detection}
We assessed whether humans could distinguish human-written false claims from AI-generated ones by annotating 50 randomly selected samples from each dataset, resulting in 150 samples in total. 12 volunteers were recruited from our NLP research group, all of whom are fluent English speakers and have some experience with AI text generation. None of the annotators had prior involvement in the research. Annotators were randomly assigned texts until all claims had received 3 annotations or the individual had annotated 90 documents (i.e. 60\% of the claims). Each claim was assigned a label based on the majority vote of the 3 annotations.

Annotations were collected with \anon[a web-based annotation tool]{GATE Teamware 2 \citep{wilby-etal-2023-gate}}. A copy of the annotation instructions and form is provided in the accompanying repository.

\begin{table*}
    \caption{Comparison of F1 performance of each method.}
    \begin{center}
    \begin{tabular}{l r r r r }
        \toprule
        \multirow{2}{*}{Detector} & \multicolumn{3}{c}{F1 of True Class} & F1-Macro \\ \cmidrule(r){2-4}\cmidrule(l){5-5}
        & ClaimReview & ChatGPT-General & ChatGPT-Topic & All \\
        \midrule
        RoBERTa & 1.00 & 0.00 & 0.00  & 0.25 \\
        ZeroGPT & 0.99 & 0.20& 0.05& 0.32\\
        Majority Annotator Vote$^*$ & 0.77 & 0.51 & 0.67 & 0.48 \\
        Random & 0.67 & 0.67 & 0.67 & 0.49 \\
        MDeBERTa & 0.19 & 1.00 & 1.00 & 0.50 \\
        \bottomrule
    \end{tabular}
    \end{center}
    \footnotesize$^*$Annotations of proportional sample of data.
    \label{tab:overall_comparison}
\end{table*}

\begin{table*}
\caption{Proportion of texts in the datasets assigned to each fine ZeroGPT result, and assigned boolean label.}
\label{tab:zerogpt_complete}
\resizebox{\textwidth}{!}{
\begin{tabular}{c p{8cm} r r r}
\toprule
Label & ZeroGPT Result                                                               & ClaimReview & ChatGPT-General & ChatGPT-Topic \\
\midrule
& Your text is Human written                                                    & 96.81\%& 89.01\%& 97.52\%\\
& Your text is Most Likely Human written                                        & 0.00\%      & 0.00\%          & 0.00\%                 \\
& Your text is Most Likely Human written, may include parts generated by AI/GPT & 0.35\%& 0.00\%& 0.00\%\\
\multirow{-4}{*}{\rotatebox[origin=c]{90}{Human}}& Your text is Likely Human written, may include parts generated by AI/GPT      & 0.00\%      & 0.00\%          & 0.00\%                 \\ \midrule
& Your text contains mixed signals, with some parts generated by AI/GPT         & 2.13\%& 10.99\%& 2.48\%\\
& Most of Your text is AI/GPT Generated                                         & 0.35\%& 0.00\%          & 0.00\%                 \\
& Your text is Likely generated by AI/GPT                                       & 0.00\%      & 0.00\%          & 0.00\%                 \\
& Your text is Most Likely AI/GPT generated                                     & 0.00\%& 0.00\%& 0.00\%\\
\multirow{-5}{*}{\rotatebox[origin=c]{90}{Machine}} & Your text is AI/GPT Generated                                                 & 0.35\%& 0.00\%          & 0.00\%  \\
\bottomrule
\end{tabular}
}
\end{table*}

\subsection{Results}
Annotator performance metrics are shown in table \ref{tab:overall_comparison}. All Fleiss' Kappa inter-annotator agreement \citep{fleiss1971measuring} values were very low or negative, indicating the agreement expected by chance or worse.

Human annotators struggled to distinguish ChatGPT-generated claims from real ones due to over-prediction of human authorship, as seen in previous work \citep{Clark-Human-Evaluation-2021f}. This causes higher F1 for the ClaimReview dataset, but poor F1-Macro overall.

Although only approximately 20\% of texts from each dataset were annotated, the extremely poor performance and alignment with previous work suggests this would also hold true were all texts annotated.

After the task, annotators reported that they felt entirely unable to distinguish the content, and were essentially guessing\footnote{We did not formally survey annotators' confidence in their ability, but prior work shows it is reduced after the task before performance is known \citep{spitale23-disinforms-better}}. As in \citet{Clark-Human-Evaluation-2021f}, annotators reported using grammatical and spelling errors to imply the claim's source may be human. It is unclear if this approach improved results on the ClaimReview set, as accuracy is only slightly higher than the proportion of claims labelled human (59\%).

\section{Computational AI Detection}

We evaluated the performance of several methods of AI content detection using our datasets.

Fine-tuned transformer models have been found to effectively detect machine-generated content on older generative models \citep{uchendu-etal-2020-authorship} and newer including ChatGPT \citep{macko-etal-2023-multitude}. We fine-tuned a \textbf{RoBERTa} \citep{liu2019roberta} model following \citeauthor{uchendu-etal-2020-authorship}, and used the best model\footnote{A variant trained on all data splits.} reported by \citeauthor{macko-etal-2023-multitude}, a fine-tuned \textbf{MDeBERTa} model \citep{he2023debertav3}. We record both the classifier's output label and accompanying certainty score.

For the RoBERTa model, we use the Balanced P2 dataset (with an equal number of texts from 9 models) provided by \citep{uchendu-etal-2020-authorship}, randomly split, stratified by label, 80\%/20\% into train and validation sets. We re-implement the training procedure using the HuggingFace trainer, as we were unable to locate training code for this model. The model was trained for 10 epochs, with a batch size of 16, learning rate of 5e-5, and the AdamW optimiser. The epoch with the highest validation F1 was selected.

An alternative method, \textbf{DetectGPT} \citep{mitchell2023detectgpt}, which uses a criterion to detect language model-generated text based on the negative curvature of the model's log probability function, was considered, however it requires a minimum of 48 tokens for reliable classification, higher than the average size of our short claims.

Finally, a commercial detection tool, \textbf{ZeroGPT}\footnote{\url{https://www.zerogpt.com/}}, was considered. It does not have a stated minimum length and was able to produce predictions on the texts.
Public information on this model is limited, but it claims to use a multi-stage method of ``complex and deep algorithms'', with a claimed accuracy of 98\%. Results are given on a categorical scale covering certainty and the possibility of mixed human/machine-generated text.

\subsection{Experimental Results}
A comparison of results for all detectors and human annotators is shown in table \ref{tab:overall_comparison}, compared to the score achieved by a random classifier.

The \textbf{RoBERTa} model identified all texts as human authored with high certainty (mean$\pm$std = $0.943 \pm 0.005$). This is likely because it is trained on older models such as GPT-2, whose lower-quality outputs are easier to identify. The \textbf{MDeBERTa} model predicted machine-authored for all of the two ChatGPT datasets and 90\% of the human-authored data, achieving an overall F1-Macro of 0.5. This is far worse than the 0.94 F1-Macro reported by \citeauthor{macko-etal-2023-multitude} for English texts, presumably due to the much shorter length of our claims. The confidence score given to human texts misclassified as machine texts was lower (0.91 $\pm$ 0.10) than correctly classified machine texts (0.98 $\pm$ 0.03), indicating there are some detectable differences, but not enough to influence the predicted label or be resolvable by adjusting the decision boundary.

\textbf{ZeroGPT} predicted human authorship for the majority of each dataset (see table \ref{tab:zerogpt_complete}). Only the weakest AI-generated label (``mixed signals'') was applied to ChatGPT-generated claims, whilst stronger AI-generated labels were assigned to a minority of ClaimReview texts.  After aggregating the label scheme to a binary human/AI label (indicated by the \emph{Label} column in table \ref{tab:zerogpt_complete}), it achieved an F1 of 0.99 on the ClaimReview data, but 0.20 and 0.05 on ChatGPT-General and -Topic respectively. The markedly worse performance on ChatGPT-Topic suggests the factors ZeroGPT uses are not consistently present in all output texts. %
The performance of ZeroGPT on these texts is significantly worse than recent results from literature, with \citet{macko-etal-2023-multitude} finding an F1-Macro on English texts of of 0.60, compared to their MDeBERTa result of 0.94.

\section{Conclusion}

This paper demonstrated the ability of ChatGPT to generate disinformation claims without providing human-authored seed narratives, including about events unknown to the model. We showed that this content cannot be reliably identified as AI-generated by humans or computational methods, that our findings hold for both general disinformation on a topic and disinformation regarding specific events, and that the model behaves differently in these two scenarios.

Our finding that ChatGPT is able to create disinformation content from scratch has wider implications for LLM generated disinformation, as previous work had only shown effective generation of complete disinformation articles from a human-authored false claim. It should be noted that as the capability and behaviour of ChatGPT changes over time \citep{chen2023chatgpts}, so will its utility for disinformation, although it has been suggested that ChatGPT was \emph{safer} when we generated our claims than subsequent versions \citep{Vykopal-DisinformationCapabilitiesModels-2023c}. 

Although our generation was limited by the number of human-authored claims found, this would not apply to a bad actor. In practical terms, this means not only can ChatGPT be used by malicious actors to propagate AI-generated disinformation and drown out reliable information on social media platforms, but that it is unlikely that the public will be able to identify the posts of such disinformation bots as AI-generated due to their fluency. 

Finally, our findings reinforce the need for transparency in commercial AI-based disinformation detection products such as ZeroGPT. It is likely its poor performance was due to text length, but this is difficult to validate given its closed source nature.

\section*{Ethics}

This research, including the human annotation of texts, was approved by an institutional ethics review process.

ChatGPT is not an open-source model, and OpenAI have previously been criticsed for their lack of transparency in their research. However, given that it is much more widely used by non-experts than open-source LLMs, we argue it is important for research on potential harms to study it.

We have chosen not to release the exact prompt used in order to minimise its misuse by bad actors to create AI-generated disinformation, particularly as we have found the prompt is still quite effective in current versions of ChatGPT. It would also be trivial to modify the prompt to generate disinformation on other topics. We describe the general strategy used by the prompt in section \ref{sec:prompt}, which we believe is sufficient for a researcher with knowledge of prompt engineering to understand the safeguard bypass-strategy we used, whilst preventing the prompt from being trivially copied. 

Likewise, we will not be publicly releasing our complete datasets. Although there is already significant disinformation surrounding the war in Ukraine, we are concerned that publicly releasing a dataset of synthetic disinformation claims would greatly aid generating long-form disinformation. Furthermore, we cannot guarantee that outputs do not reproduce content which is under copyright.

We will provide the complete prompts and datasets in a restricted Zenodo repository, which will be made available for research purposes only upon request. This approach is in-line with other works pertaining to disinformation or LLM outputs, e.g. \citet{macko-etal-2023-multitude}.

To the best of our knowledge, the generated claims do not ever directly or indirectly mention individuals.  None of the generated claims contain content which is inherently offensive (e.g. abusive or disturbing content), but as it is disinformation surrounding a sensitive topic, some may consider it offensive nonetheless.

Human annotation data is reported as three aggregated and anonymous votes, with no possibility to identify any individual annotator.

\begin{acks}
This work is partially supported by \grantsponsor{InnovateUK}{InnovateUK}{https://www.ukri.org/councils/innovate-uk/} grant number \grantnum{InnovateUK}{10039055} (approved under \grantsponsor{HorizonEurope}{Horizon Europe}{https://research-and-innovation.ec.europa.eu/funding/funding-opportunities/funding-programmes-and-open-calls/horizon-europe_en} as Vera.ai\footnote{\url{https://www.veraai.eu/}}, grant agreement number \grantnum{HorizonEurope}{101070093}). Freddy Heppell is supported by a \grantsponsor{UoSEng}{University of Sheffield Faculty of Engineering}{https://www.sheffield.ac.uk/engineering} \grantnum{UosEng}{PGR Prize Scholarship}.

The authors also want to thank Ivan Srba and the KInIT team for providing us with access to their best MDeBERTa detection model.
\end{acks}



\begin{thebibliography}{62}


\ifx \showCODEN    \undefined \def \showCODEN     #1{\unskip}     \fi
\ifx \showDOI      \undefined \def \showDOI       #1{#1}\fi
\ifx \showISBNx    \undefined \def \showISBNx     #1{\unskip}     \fi
\ifx \showISBNxiii \undefined \def \showISBNxiii  #1{\unskip}     \fi
\ifx \showISSN     \undefined \def \showISSN      #1{\unskip}     \fi
\ifx \showLCCN     \undefined \def \showLCCN      #1{\unskip}     \fi
\ifx \shownote     \undefined \def \shownote      #1{#1}          \fi
\ifx \showarticletitle \undefined \def \showarticletitle #1{#1}   \fi
\ifx \showURL      \undefined \def \showURL       {\relax}        \fi
\providecommand\bibfield[2]{#2}
\providecommand\bibinfo[2]{#2}
\providecommand\natexlab[1]{#1}
\providecommand\showeprint[2][]{arXiv:#2}

\bibitem[Augenstein et~al\mbox{.}(2023)]%
        {augenstein2023factualitychallengeseralarge}
\bibfield{author}{\bibinfo{person}{Isabelle Augenstein}, \bibinfo{person}{Timothy Baldwin}, \bibinfo{person}{Meeyoung Cha}, \bibinfo{person}{Tanmoy Chakraborty}, \bibinfo{person}{Giovanni~Luca Ciampaglia}, \bibinfo{person}{David Corney}, \bibinfo{person}{Renee DiResta}, \bibinfo{person}{Emilio Ferrara}, \bibinfo{person}{Scott Hale}, \bibinfo{person}{Alon Halevy}, \bibinfo{person}{Eduard Hovy}, \bibinfo{person}{Heng Ji}, \bibinfo{person}{Filippo Menczer}, \bibinfo{person}{Ruben Miguez}, \bibinfo{person}{Preslav Nakov}, \bibinfo{person}{Dietram Scheufele}, \bibinfo{person}{Shivam Sharma}, {and} \bibinfo{person}{Giovanni Zagni}.} \bibinfo{year}{2023}\natexlab{}.
\newblock \bibinfo{title}{Factuality Challenges in the Era of Large Language Models}.
\newblock
\newblock
\showeprint[arxiv]{2310.05189}~[cs.CL]


\bibitem[Bakhtin et~al\mbox{.}(2019)]%
        {Bakhtin-RealFakeText-2019r}
\bibfield{author}{\bibinfo{person}{Anton Bakhtin}, \bibinfo{person}{Sam Gross}, \bibinfo{person}{Myle Ott}, \bibinfo{person}{Yuntian Deng}, \bibinfo{person}{Marc'Aurelio Ranzato}, {and} \bibinfo{person}{Arthur Szlam}.} \bibinfo{year}{2019}\natexlab{}.
\newblock \bibinfo{title}{Real or Fake? Learning to Discriminate Machine from Human Generated Text}.
\newblock
\newblock
\showeprint[arxiv]{1906.03351}~[cs.LG]


\bibitem[Bashardoust et~al\mbox{.}(2024)]%
        {Bashardoust-ComparingWillingnessNews-2024f}
\bibfield{author}{\bibinfo{person}{Amirsiavosh Bashardoust}, \bibinfo{person}{Stefan Feuerriegel}, {and} \bibinfo{person}{Yash~Raj Shrestha}.} \bibinfo{year}{2024}\natexlab{}.
\newblock \bibinfo{title}{Comparing the willingness to share for human-generated vs. {AI}-generated fake news}.
\newblock
\newblock
\showeprint[arxiv]{2402.07395}~[cs.SI]
\urldef\tempurl%
\url{http://arxiv.org/abs/2402.07395}
\showURL{%
\tempurl}


\bibitem[Boyd et~al\mbox{.}(2022)]%
        {liwc22dev}
\bibfield{author}{\bibinfo{person}{Ryan~L Boyd}, \bibinfo{person}{Ashwini Ashokkumar}, \bibinfo{person}{Sarah Seraj}, {and} \bibinfo{person}{James~W Pennebaker}.} \bibinfo{year}{2022}\natexlab{}.
\newblock \bibinfo{title}{The Development and Psychometric Properties of {LIWC-22}}.
\newblock
\newblock
\urldef\tempurl%
\url{https://www.liwc.app/static/documents/LIWC-22%20Manual%20-%20Development%20and%20Psychometrics.pdf}
\showURL{%
\tempurl}


\bibitem[Brown et~al\mbox{.}(2020)]%
        {brown2020language}
\bibfield{author}{\bibinfo{person}{Tom Brown}, \bibinfo{person}{Benjamin Mann}, \bibinfo{person}{Nick Ryder}, \bibinfo{person}{Melanie Subbiah}, \bibinfo{person}{Jared~D Kaplan}, \bibinfo{person}{Prafulla Dhariwal}, \bibinfo{person}{Arvind Neelakantan}, \bibinfo{person}{Pranav Shyam}, \bibinfo{person}{Girish Sastry}, \bibinfo{person}{Amanda Askell}, \bibinfo{person}{Sandhini Agarwal}, \bibinfo{person}{Ariel Herbert-Voss}, \bibinfo{person}{Gretchen Krueger}, \bibinfo{person}{Tom Henighan}, \bibinfo{person}{Rewon Child}, \bibinfo{person}{Aditya Ramesh}, \bibinfo{person}{Daniel Ziegler}, \bibinfo{person}{Jeffrey Wu}, \bibinfo{person}{Clemens Winter}, \bibinfo{person}{Chris Hesse}, \bibinfo{person}{Mark Chen}, \bibinfo{person}{Eric Sigler}, \bibinfo{person}{Mateusz Litwin}, \bibinfo{person}{Scott Gray}, \bibinfo{person}{Benjamin Chess}, \bibinfo{person}{Jack Clark}, \bibinfo{person}{Christopher Berner}, \bibinfo{person}{Sam McCandlish}, \bibinfo{person}{Alec Radford}, \bibinfo{person}{Ilya Sutskever}, {and}
  \bibinfo{person}{Dario Amodei}.} \bibinfo{year}{2020}\natexlab{}.
\newblock \showarticletitle{Language Models are Few-Shot Learners}. In \bibinfo{booktitle}{\emph{Proceedings of the 34th International Conference on Neural Information Processing Systems}} (Online), \bibfield{editor}{\bibinfo{person}{H.~Larochelle}, \bibinfo{person}{M.~Ranzato}, \bibinfo{person}{R.~Hadsell}, \bibinfo{person}{M.F. Balcan}, {and} \bibinfo{person}{H.~Lin}} (Eds.), Vol.~\bibinfo{volume}{33}. \bibinfo{publisher}{Curran Associates, Inc.}, \bibinfo{address}{Red Hook, NY, USA}, \bibinfo{pages}{1877--1901}.
\newblock
\urldef\tempurl%
\url{https://proceedings.neurips.cc/paper_files/paper/2020/file/1457c0d6bfcb4967418bfb8ac142f64a-Paper.pdf}
\showURL{%
\tempurl}


\bibitem[Buchanan et~al\mbox{.}(2021)]%
        {Buchanan-TruthLiesDisinformation-2021d}
\bibfield{author}{\bibinfo{person}{Ben Buchanan}, \bibinfo{person}{Andrew Lohn}, \bibinfo{person}{Micah Musser}, {and} \bibinfo{person}{Katerina Sedova}.} \bibinfo{year}{2021}\natexlab{}.
\newblock \bibinfo{booktitle}{\emph{Truth, lies, and automation how language models could change disinformation}}.
\newblock \bibinfo{type}{{T}echnical {R}eport}. \bibinfo{institution}{Centre for Security and Emerging Technology}.
\newblock
\urldef\tempurl%
\url{https://cset.georgetown.edu/wp-content/uploads/CSET-Truth-Lies-and-Automation.pdf}
\showURL{%
\tempurl}


\bibitem[Busker et~al\mbox{.}(2023)]%
        {10.1145/3614321.3614325}
\bibfield{author}{\bibinfo{person}{Tony Busker}, \bibinfo{person}{Sunil Choenni}, {and} \bibinfo{person}{Mortaza Shoae~Bargh}.} \bibinfo{year}{2023}\natexlab{}.
\newblock \showarticletitle{Stereotypes in ChatGPT: an empirical study}. In \bibinfo{booktitle}{\emph{Proceedings of the 16th International Conference on Theory and Practice of Electronic Governance}} (Belo Horizonte, Brazil) \emph{(\bibinfo{series}{ICEGOV '23})}. \bibinfo{publisher}{Association for Computing Machinery}, \bibinfo{address}{New York, NY, USA}, \bibinfo{pages}{24–32}.
\newblock
\showISBNx{9798400707421}
\urldef\tempurl%
\url{https://doi.org/10.1145/3614321.3614325}
\showDOI{\tempurl}


\bibitem[Cai and Cui(2023)]%
        {cai2023evadechatgptdetectorssingle}
\bibfield{author}{\bibinfo{person}{Shuyang Cai} {and} \bibinfo{person}{Wanyun Cui}.} \bibinfo{year}{2023}\natexlab{}.
\newblock \bibinfo{title}{Evade ChatGPT Detectors via A Single Space}.
\newblock
\newblock
\showeprint[arxiv]{2307.02599}~[cs.CL]


\bibitem[Chen and Shu(2023)]%
        {chen2023combating}
\bibfield{author}{\bibinfo{person}{Canyu Chen} {and} \bibinfo{person}{Kai Shu}.} \bibinfo{year}{2023}\natexlab{}.
\newblock \bibinfo{title}{Combating Misinformation in the Age of LLMs: Opportunities and Challenges}.
\newblock
\newblock
\showeprint[arxiv]{2311.05656}~[cs.CY]


\bibitem[Chen and Shu(2024)]%
        {Chen-LLM-generatedMisinformation-2023o}
\bibfield{author}{\bibinfo{person}{Canyu Chen} {and} \bibinfo{person}{Kai Shu}.} \bibinfo{year}{2024}\natexlab{}.
\newblock \bibinfo{title}{Can LLM-Generated Misinformation Be Detected?}
\newblock
\newblock
\showeprint[arxiv]{2309.13788}~[cs.CL]


\bibitem[Chen and Zong(2003)]%
        {tfsqrt}
\bibfield{author}{\bibinfo{person}{Keli Chen} {and} \bibinfo{person}{Chengqing Zong}.} \bibinfo{year}{2003}\natexlab{}.
\newblock \showarticletitle{A new weighting algorithm for linear classifier}. In \bibinfo{booktitle}{\emph{International Conference on Natural Language Processing and Knowledge Engineering, 2003. Proceedings. 2003}} (Beijing, China). \bibinfo{pages}{650--655}.
\newblock
\urldef\tempurl%
\url{https://doi.org/10.1109/NLPKE.2003.1275987}
\showDOI{\tempurl}


\bibitem[Chen et~al\mbox{.}(2023)]%
        {chen2023chatgpts}
\bibfield{author}{\bibinfo{person}{Lingjiao Chen}, \bibinfo{person}{Matei Zaharia}, {and} \bibinfo{person}{James Zou}.} \bibinfo{year}{2023}\natexlab{}.
\newblock \bibinfo{title}{How is ChatGPT's behavior changing over time?}
\newblock
\newblock
\showeprint[arxiv]{2307.09009}~[cs.CL]


\bibitem[Cheng et~al\mbox{.}(2024)]%
        {cheng2024dateddatatracingknowledge}
\bibfield{author}{\bibinfo{person}{Jeffrey Cheng}, \bibinfo{person}{Marc Marone}, \bibinfo{person}{Orion Weller}, \bibinfo{person}{Dawn Lawrie}, \bibinfo{person}{Daniel Khashabi}, {and} \bibinfo{person}{Benjamin~Van Durme}.} \bibinfo{year}{2024}\natexlab{}.
\newblock \bibinfo{title}{Dated Data: Tracing Knowledge Cutoffs in Large Language Models}.
\newblock
\newblock
\showeprint[arxiv]{2403.12958}~[cs.CL]


\bibitem[Clark et~al\mbox{.}(2021)]%
        {Clark-Human-Evaluation-2021f}
\bibfield{author}{\bibinfo{person}{Elizabeth Clark}, \bibinfo{person}{Tal August}, \bibinfo{person}{Sofia Serrano}, \bibinfo{person}{Nikita Haduong}, \bibinfo{person}{Suchin Gururangan}, {and} \bibinfo{person}{Noah~A Smith}.} \bibinfo{year}{2021}\natexlab{}.
\newblock \showarticletitle{All that's `human' is not gold: Evaluating human evaluation of generated text}. In \bibinfo{booktitle}{\emph{Proceedings of the 59th Annual Meeting of the Association for Computational Linguistics and the 11th International Joint Conference on Natural Language Processing (Volume 1: Long Papers)}}. \bibinfo{publisher}{Association for Computational Linguistics}, \bibinfo{address}{Stroudsburg, PA, USA}, \bibinfo{pages}{7282--7296}.
\newblock
\urldef\tempurl%
\url{https://doi.org/10.18653/v1/2021.acl-long.565}
\showDOI{\tempurl}


\bibitem[Crothers et~al\mbox{.}(2023)]%
        {Crothers-Machine-generatedTextMethods-2023n}
\bibfield{author}{\bibinfo{person}{Evan~N Crothers}, \bibinfo{person}{Nathalie Japkowicz}, {and} \bibinfo{person}{Herna~L Viktor}.} \bibinfo{year}{2023}\natexlab{}.
\newblock \showarticletitle{Machine-generated text: A comprehensive survey of threat models and detection methods}.
\newblock \bibinfo{journal}{\emph{IEEE Access}}  \bibinfo{volume}{11} (\bibinfo{year}{2023}), \bibinfo{pages}{70977--71002}.
\newblock
\showISSN{2169-3536}
\urldef\tempurl%
\url{https://doi.org/10.1109/access.2023.3294090}
\showDOI{\tempurl}


\bibitem[Fallis(2015)]%
        {Fallis-WhatDisinformation-2015p}
\bibfield{author}{\bibinfo{person}{Don Fallis}.} \bibinfo{year}{2015}\natexlab{}.
\newblock \showarticletitle{What is disinformation?}
\newblock \bibinfo{journal}{\emph{Libr. Trends}} \bibinfo{volume}{63}, \bibinfo{number}{3} (\bibinfo{year}{2015}), \bibinfo{pages}{401--426}.
\newblock
\showISSN{0024-2594,1559-0682}
\urldef\tempurl%
\url{https://doi.org/10.1353/lib.2015.0014}
\showDOI{\tempurl}


\bibitem[Fleiss(1971)]%
        {fleiss1971measuring}
\bibfield{author}{\bibinfo{person}{Joseph~L Fleiss}.} \bibinfo{year}{1971}\natexlab{}.
\newblock \showarticletitle{Measuring nominal scale agreement among many raters.}
\newblock \bibinfo{journal}{\emph{Psychological bulletin}} \bibinfo{volume}{76}, \bibinfo{number}{5} (\bibinfo{year}{1971}), \bibinfo{pages}{378}.
\newblock
\urldef\tempurl%
\url{https://doi.org/10.1037/h0031619}
\showDOI{\tempurl}


\bibitem[Goldstein et~al\mbox{.}(2023)]%
        {goldstein2023generativelanguagemodelsautomated}
\bibfield{author}{\bibinfo{person}{Josh~A. Goldstein}, \bibinfo{person}{Girish Sastry}, \bibinfo{person}{Micah Musser}, \bibinfo{person}{Renee DiResta}, \bibinfo{person}{Matthew Gentzel}, {and} \bibinfo{person}{Katerina Sedova}.} \bibinfo{year}{2023}\natexlab{}.
\newblock \bibinfo{title}{Generative Language Models and Automated Influence Operations: Emerging Threats and Potential Mitigations}.
\newblock
\newblock
\showeprint[arxiv]{2301.04246}~[cs.CY]


\bibitem[Grootendorst(2022)]%
        {grootendorst2022bertopic}
\bibfield{author}{\bibinfo{person}{Maarten Grootendorst}.} \bibinfo{year}{2022}\natexlab{}.
\newblock \bibinfo{title}{BERTopic: Neural topic modeling with a class-based TF-IDF procedure}.
\newblock
\newblock
\showeprint[arxiv]{2203.05794}~[cs.CL]


\bibitem[He et~al\mbox{.}(2023)]%
        {he2023debertav3}
\bibfield{author}{\bibinfo{person}{Pengcheng He}, \bibinfo{person}{Jianfeng Gao}, {and} \bibinfo{person}{Weizhu Chen}.} \bibinfo{year}{2023}\natexlab{}.
\newblock \bibinfo{title}{DeBERTaV3: Improving DeBERTa using ELECTRA-Style Pre-Training with Gradient-Disentangled Embedding Sharing}.
\newblock
\newblock
\showeprint[arxiv]{2111.09543}~[cs.CL]


\bibitem[Huang and Sun(2023)]%
        {Huang-HarnessingPowerExplanation-2023u}
\bibfield{author}{\bibinfo{person}{Yue Huang} {and} \bibinfo{person}{Lichao Sun}.} \bibinfo{year}{2023}\natexlab{}.
\newblock \bibinfo{title}{Harnessing the Power of {ChatGPT} in Fake News: An In-Depth Exploration in Generation, Detection and Explanation}.
\newblock
\newblock
\showeprint[arxiv]{2310.05046}~[cs.CL]


\bibitem[Ippolito et~al\mbox{.}(2020)]%
        {ippolito-etal-2020-automatic}
\bibfield{author}{\bibinfo{person}{Daphne Ippolito}, \bibinfo{person}{Daniel Duckworth}, \bibinfo{person}{Chris Callison-Burch}, {and} \bibinfo{person}{Douglas Eck}.} \bibinfo{year}{2020}\natexlab{}.
\newblock \showarticletitle{Automatic Detection of Generated Text is Easiest when Humans are Fooled}. In \bibinfo{booktitle}{\emph{Proceedings of the 58th Annual Meeting of the Association for Computational Linguistics}}, \bibfield{editor}{\bibinfo{person}{Dan Jurafsky}, \bibinfo{person}{Joyce Chai}, \bibinfo{person}{Natalie Schluter}, {and} \bibinfo{person}{Joel Tetreault}} (Eds.). \bibinfo{publisher}{Association for Computational Linguistics}, \bibinfo{address}{Online}, \bibinfo{pages}{1808--1822}.
\newblock
\urldef\tempurl%
\url{https://doi.org/10.18653/v1/2020.acl-main.164}
\showDOI{\tempurl}


\bibitem[Jiang et~al\mbox{.}(2024)]%
        {jiang-etal-2024-large}
\bibfield{author}{\bibinfo{person}{Che Jiang}, \bibinfo{person}{Biqing Qi}, \bibinfo{person}{Xiangyu Hong}, \bibinfo{person}{Dayuan Fu}, \bibinfo{person}{Yang Cheng}, \bibinfo{person}{Fandong Meng}, \bibinfo{person}{Mo Yu}, \bibinfo{person}{Bowen Zhou}, {and} \bibinfo{person}{Jie Zhou}.} \bibinfo{year}{2024}\natexlab{}.
\newblock \showarticletitle{On Large Language Models{'} Hallucination with Regard to Known Facts}. In \bibinfo{booktitle}{\emph{Proceedings of the 2024 Conference of the North American Chapter of the Association for Computational Linguistics: Human Language Technologies (Volume 1: Long Papers)}}, \bibfield{editor}{\bibinfo{person}{Kevin Duh}, \bibinfo{person}{Helena Gomez}, {and} \bibinfo{person}{Steven Bethard}} (Eds.). \bibinfo{publisher}{Association for Computational Linguistics}, \bibinfo{address}{Mexico City, Mexico}, \bibinfo{pages}{1041--1053}.
\newblock
\urldef\tempurl%
\url{https://aclanthology.org/2024.naacl-long.60}
\showURL{%
\tempurl}


\bibitem[Joulin et~al\mbox{.}(2016)]%
        {joulin2016fasttextzip}
\bibfield{author}{\bibinfo{person}{Armand Joulin}, \bibinfo{person}{Edouard Grave}, \bibinfo{person}{Piotr Bojanowski}, \bibinfo{person}{Matthijs Douze}, \bibinfo{person}{Hérve Jégou}, {and} \bibinfo{person}{Tomas Mikolov}.} \bibinfo{year}{2016}\natexlab{}.
\newblock \bibinfo{title}{FastText.zip: Compressing text classification models}.
\newblock
\newblock
\showeprint[arxiv]{1612.03651}~[cs.CL]


\bibitem[Joulin et~al\mbox{.}(2017)]%
        {joulin-etal-2017-bag}
\bibfield{author}{\bibinfo{person}{Armand Joulin}, \bibinfo{person}{Edouard Grave}, \bibinfo{person}{Piotr Bojanowski}, {and} \bibinfo{person}{Tomas Mikolov}.} \bibinfo{year}{2017}\natexlab{}.
\newblock \showarticletitle{Bag of Tricks for Efficient Text Classification}. In \bibinfo{booktitle}{\emph{Proceedings of the 15th Conference of the {E}uropean Chapter of the Association for Computational Linguistics: Volume 2, Short Papers}}, \bibfield{editor}{\bibinfo{person}{Mirella Lapata}, \bibinfo{person}{Phil Blunsom}, {and} \bibinfo{person}{Alexander Koller}} (Eds.). \bibinfo{publisher}{Association for Computational Linguistics}, \bibinfo{address}{Valencia, Spain}, \bibinfo{pages}{427--431}.
\newblock
\urldef\tempurl%
\url{https://aclanthology.org/E17-2068}
\showURL{%
\tempurl}


\bibitem[Kirchenbauer et~al\mbox{.}(2023)]%
        {pmlr-v202-kirchenbauer23a}
\bibfield{author}{\bibinfo{person}{John Kirchenbauer}, \bibinfo{person}{Jonas Geiping}, \bibinfo{person}{Yuxin Wen}, \bibinfo{person}{Jonathan Katz}, \bibinfo{person}{Ian Miers}, {and} \bibinfo{person}{Tom Goldstein}.} \bibinfo{year}{2023}\natexlab{}.
\newblock \showarticletitle{A Watermark for Large Language Models}. In \bibinfo{booktitle}{\emph{Proceedings of the 40th International Conference on Machine Learning}} \emph{(\bibinfo{series}{Proceedings of Machine Learning Research}, Vol.~\bibinfo{volume}{202})}, \bibfield{editor}{\bibinfo{person}{Andreas Krause}, \bibinfo{person}{Emma Brunskill}, \bibinfo{person}{Kyunghyun Cho}, \bibinfo{person}{Barbara Engelhardt}, \bibinfo{person}{Sivan Sabato}, {and} \bibinfo{person}{Jonathan Scarlett}} (Eds.). \bibinfo{publisher}{PMLR}, \bibinfo{pages}{17061--17084}.
\newblock
\urldef\tempurl%
\url{https://proceedings.mlr.press/v202/kirchenbauer23a.html}
\showURL{%
\tempurl}


\bibitem[Kornbrot(2014)]%
        {pbcorr}
\bibfield{author}{\bibinfo{person}{Diana Kornbrot}.} \bibinfo{year}{2014}\natexlab{}.
\newblock \showarticletitle{Point Biserial Correlation}.
\newblock In \bibinfo{booktitle}{\emph{Wiley StatsRef: Statistics Reference Online}}. \bibinfo{publisher}{John Wiley \& Sons, Ltd}.
\newblock
\showISBNx{9781118445112}
\urldef\tempurl%
\url{https://doi.org/10.1002/9781118445112.stat06227}
\showDOI{\tempurl}


\bibitem[Kreps et~al\mbox{.}(2022)]%
        {Kreps-NewsThatsMisinformation-2022l}
\bibfield{author}{\bibinfo{person}{Sarah Kreps}, \bibinfo{person}{R.~Miles McCain}, {and} \bibinfo{person}{Miles Brundage}.} \bibinfo{year}{2022}\natexlab{}.
\newblock \showarticletitle{All the news that’s fit to fabricate: {AI}-generated text as a tool of media misinformation}.
\newblock \bibinfo{journal}{\emph{J. Exp. Polit. Sci.}} \bibinfo{volume}{9}, \bibinfo{number}{1} (\bibinfo{year}{2022}), \bibinfo{pages}{104--117}.
\newblock
\showISSN{2052-2630,2052-2649}
\urldef\tempurl%
\url{https://doi.org/10.1017/xps.2020.37}
\showDOI{\tempurl}


\bibitem[Lee et~al\mbox{.}(2023)]%
        {Lee2023-vp}
\bibfield{author}{\bibinfo{person}{Sian Lee}, \bibinfo{person}{Aiping Xiong}, \bibinfo{person}{Haeseung Seo}, {and} \bibinfo{person}{Dongwon Lee}.} \bibinfo{year}{2023}\natexlab{}.
\newblock \showarticletitle{``Fact-checking'' fact checkers: A data-driven approach}.
\newblock \bibinfo{journal}{\emph{Harvard Kennedy School (HKS) Misinformation Review}} (\bibinfo{date}{Oct.} \bibinfo{year}{2023}).
\newblock
\urldef\tempurl%
\url{https://doi.org/10.37016/mr-2020-126}
\showDOI{\tempurl}


\bibitem[Li et~al\mbox{.}(2024)]%
        {li2024chatgptdoesnttrustchargers}
\bibfield{author}{\bibinfo{person}{Victoria~R. Li}, \bibinfo{person}{Yida Chen}, {and} \bibinfo{person}{Naomi Saphra}.} \bibinfo{year}{2024}\natexlab{}.
\newblock \bibinfo{title}{ChatGPT Doesn't Trust Chargers Fans: Guardrail Sensitivity in Context}.
\newblock
\newblock
\showeprint[arxiv]{2407.06866}~[cs.CL]
\urldef\tempurl%
\url{https://arxiv.org/abs/2407.06866}
\showURL{%
\tempurl}


\bibitem[Liang et~al\mbox{.}(2023)]%
        {2023nonnativewriters}
\bibfield{author}{\bibinfo{person}{Weixin Liang}, \bibinfo{person}{Mert Yuksekgonul}, \bibinfo{person}{Yining Mao}, \bibinfo{person}{Eric Wu}, {and} \bibinfo{person}{James Zou}.} \bibinfo{year}{2023}\natexlab{}.
\newblock \showarticletitle{GPT detectors are biased against non-native English writers}.
\newblock \bibinfo{journal}{\emph{Patterns}} \bibinfo{volume}{4}, \bibinfo{number}{7} (\bibinfo{year}{2023}), \bibinfo{pages}{100779}.
\newblock
\showISSN{2666-3899}
\urldef\tempurl%
\url{https://doi.org/10.1016/j.patter.2023.100779}
\showDOI{\tempurl}


\bibitem[Liu et~al\mbox{.}(2019)]%
        {liu2019roberta}
\bibfield{author}{\bibinfo{person}{Yinhan Liu}, \bibinfo{person}{Myle Ott}, \bibinfo{person}{Naman Goyal}, \bibinfo{person}{Jingfei Du}, \bibinfo{person}{Mandar Joshi}, \bibinfo{person}{Danqi Chen}, \bibinfo{person}{Omer Levy}, \bibinfo{person}{Mike Lewis}, \bibinfo{person}{Luke Zettlemoyer}, {and} \bibinfo{person}{Veselin Stoyanov}.} \bibinfo{year}{2019}\natexlab{}.
\newblock \bibinfo{title}{RoBERTa: A Robustly Optimized BERT Pretraining Approach}.
\newblock
\newblock
\showeprint[arxiv]{1907.11692}~[cs.CL]


\bibitem[Lucas et~al\mbox{.}(2023)]%
        {lucas-etal-2023-fighting}
\bibfield{author}{\bibinfo{person}{Jason Lucas}, \bibinfo{person}{Adaku Uchendu}, \bibinfo{person}{Michiharu Yamashita}, \bibinfo{person}{Jooyoung Lee}, \bibinfo{person}{Shaurya Rohatgi}, {and} \bibinfo{person}{Dongwon Lee}.} \bibinfo{year}{2023}\natexlab{}.
\newblock \showarticletitle{Fighting Fire with Fire: The Dual Role of {LLM}s in Crafting and Detecting Elusive Disinformation}. In \bibinfo{booktitle}{\emph{Proceedings of the 2023 Conference on Empirical Methods in Natural Language Processing}}, \bibfield{editor}{\bibinfo{person}{Houda Bouamor}, \bibinfo{person}{Juan Pino}, {and} \bibinfo{person}{Kalika Bali}} (Eds.). \bibinfo{publisher}{Association for Computational Linguistics}, \bibinfo{address}{Singapore}, \bibinfo{pages}{14279--14305}.
\newblock
\urldef\tempurl%
\url{https://doi.org/10.18653/v1/2023.emnlp-main.883}
\showDOI{\tempurl}


\bibitem[Macko et~al\mbox{.}(2023)]%
        {macko-etal-2023-multitude}
\bibfield{author}{\bibinfo{person}{Dominik Macko}, \bibinfo{person}{Robert Moro}, \bibinfo{person}{Adaku Uchendu}, \bibinfo{person}{Jason Lucas}, \bibinfo{person}{Michiharu Yamashita}, \bibinfo{person}{Mat{\'u}{\v{s}} Pikuliak}, \bibinfo{person}{Ivan Srba}, \bibinfo{person}{Thai Le}, \bibinfo{person}{Dongwon Lee}, \bibinfo{person}{Jakub Simko}, {and} \bibinfo{person}{Maria Bielikova}.} \bibinfo{year}{2023}\natexlab{}.
\newblock \showarticletitle{{MULTIT}u{DE}: Large-Scale Multilingual Machine-Generated Text Detection Benchmark}. In \bibinfo{booktitle}{\emph{Proceedings of the 2023 Conference on Empirical Methods in Natural Language Processing}}, \bibfield{editor}{\bibinfo{person}{Houda Bouamor}, \bibinfo{person}{Juan Pino}, {and} \bibinfo{person}{Kalika Bali}} (Eds.). \bibinfo{publisher}{Association for Computational Linguistics}, \bibinfo{address}{Singapore}, \bibinfo{pages}{9960--9987}.
\newblock
\urldef\tempurl%
\url{https://doi.org/10.18653/v1/2023.emnlp-main.616}
\showDOI{\tempurl}


\bibitem[Macko et~al\mbox{.}(2024)]%
        {macko2024authorshipobfuscation}
\bibfield{author}{\bibinfo{person}{Dominik Macko}, \bibinfo{person}{Robert Moro}, \bibinfo{person}{Adaku Uchendu}, \bibinfo{person}{Ivan Srba}, \bibinfo{person}{Jason~Samuel Lucas}, \bibinfo{person}{Michiharu Yamashita}, \bibinfo{person}{Nafis~Irtiza Tripto}, \bibinfo{person}{Dongwon Lee}, \bibinfo{person}{Jakub Simko}, {and} \bibinfo{person}{Maria Bielikova}.} \bibinfo{year}{2024}\natexlab{}.
\newblock \bibinfo{title}{Authorship Obfuscation in Multilingual Machine-Generated Text Detection}.
\newblock
\newblock
\showeprint[arxiv]{2401.07867}~[cs.CL]
\urldef\tempurl%
\url{https://arxiv.org/abs/2401.07867}
\showURL{%
\tempurl}


\bibitem[McGuffie and Newhouse(2020)]%
        {McGuffie-RadicalizationRisksModels-2020p}
\bibfield{author}{\bibinfo{person}{Kris McGuffie} {and} \bibinfo{person}{Alex Newhouse}.} \bibinfo{year}{2020}\natexlab{}.
\newblock \bibinfo{title}{The Radicalization Risks of GPT-3 and Advanced Neural Language Models}.
\newblock
\newblock
\showeprint[arxiv]{2009.06807}~[cs.CY]


\bibitem[Mitchell et~al\mbox{.}(2023)]%
        {mitchell2023detectgpt}
\bibfield{author}{\bibinfo{person}{Eric Mitchell}, \bibinfo{person}{Yoonho Lee}, \bibinfo{person}{Alexander Khazatsky}, \bibinfo{person}{Christopher~D. Manning}, {and} \bibinfo{person}{Chelsea Finn}.} \bibinfo{year}{2023}\natexlab{}.
\newblock \showarticletitle{DetectGPT: Zero-Shot Machine-Generated Text Detection Using Probability Curvature}. In \bibinfo{booktitle}{\emph{Proceedings of the 40th International Conference on Machine Learning}} \emph{(\bibinfo{series}{ICML'23})}. \bibinfo{publisher}{JMLR.org}, \bibinfo{address}{Honolulu, HI, USA}, Article \bibinfo{articleno}{1038}, \bibinfo{numpages}{13}~pages.
\newblock
\urldef\tempurl%
\url{https://proceedings.mlr.press/v202/mitchell23a.html}
\showURL{%
\tempurl}


\bibitem[Mittelstadt et~al\mbox{.}(2023)]%
        {mittelstadt2023protect}
\bibfield{author}{\bibinfo{person}{Brent Mittelstadt}, \bibinfo{person}{Sandra Wachter}, {and} \bibinfo{person}{Chris Russell}.} \bibinfo{year}{2023}\natexlab{}.
\newblock \showarticletitle{To protect science, we must use LLMs as zero-shot translators}.
\newblock \bibinfo{journal}{\emph{Nature Human Behaviour}} \bibinfo{volume}{7}, \bibinfo{number}{11} (\bibinfo{year}{2023}), \bibinfo{pages}{1830--1832}.
\newblock
\urldef\tempurl%
\url{https://doi.org/10.1038/s41562-023-01744-0}
\showDOI{\tempurl}


\bibitem[Mosallanezhad et~al\mbox{.}(2020)]%
        {Mosallanezhad-Topic-preservingSyntheticApproach-2020u}
\bibfield{author}{\bibinfo{person}{Ahmadreza Mosallanezhad}, \bibinfo{person}{Kai Shu}, {and} \bibinfo{person}{Huan Liu}.} \bibinfo{year}{2020}\natexlab{}.
\newblock \bibinfo{title}{Topic-Preserving Synthetic News Generation: An Adversarial Deep Reinforcement Learning Approach}.
\newblock
\newblock
\showeprint[arxiv]{2010.16324}~[cs.CL]


\bibitem[{OpenAI}(2023)]%
        {OpenAI-GPT-4TechnicalReport-2023i}
\bibfield{author}{\bibinfo{person}{{OpenAI}}.} \bibinfo{year}{2023}\natexlab{}.
\newblock \bibinfo{title}{{GPT}-4 Technical Report}.
\newblock
\newblock
\urldef\tempurl%
\url{https://cdn.openai.com/papers/gpt-4.pdf}
\showURL{%
\tempurl}


\bibitem[Paul and Matthews(2016)]%
        {Paul-RussianFirehoseCounter-2016g}
\bibfield{author}{\bibinfo{person}{Christopher Paul} {and} \bibinfo{person}{Miriam Matthews}.} \bibinfo{year}{2016}\natexlab{}.
\newblock \bibinfo{booktitle}{\emph{The Russian ``firehose of falsehood'' propaganda model: Why it might work and options to counter it}}.
\newblock \bibinfo{type}{{T}echnical {R}eport}. \bibinfo{institution}{RAND Corporation}.
\newblock
\urldef\tempurl%
\url{https://doi.org/10.7249/PE198}
\showDOI{\tempurl}


\bibitem[Pennebaker et~al\mbox{.}(2022)]%
        {liwc22}
\bibfield{author}{\bibinfo{person}{James~W Pennebaker}, \bibinfo{person}{Ryan~L Boyd}, \bibinfo{person}{Roger~J Booth}, \bibinfo{person}{Ashwini Ashokkumar}, {and} \bibinfo{person}{Francis~Martha E}.} \bibinfo{year}{2022}\natexlab{}.
\newblock \bibinfo{booktitle}{\emph{Linguistic Inquiry and Word Count: {LIWC}-22}}.
\newblock Software. Pennebaker Conglomerates.
\newblock
\urldef\tempurl%
\url{https://www.liwc.app/}
\showURL{%
\tempurl}


\bibitem[Radford et~al\mbox{.}(2019)]%
        {openaibettermodels}
\bibfield{author}{\bibinfo{person}{Alec Radford}, \bibinfo{person}{Jeffrey Wu}, \bibinfo{person}{Dario Amodei}, \bibinfo{person}{Jack Clark}, \bibinfo{person}{Miles Brundage}, {and} \bibinfo{person}{Ilya Sutskever}.} \bibinfo{year}{2019}\natexlab{}.
\newblock \bibinfo{booktitle}{\emph{Better language models and their implications}}.
\newblock OpenAI.
\newblock
\urldef\tempurl%
\url{https://openai.com/research/better-language-models}
\showURL{%
Retrieved April 15, 2024 from \tempurl}


\bibitem[Reimers and Gurevych(2019)]%
        {reimers-gurevych-2019-sentence}
\bibfield{author}{\bibinfo{person}{Nils Reimers} {and} \bibinfo{person}{Iryna Gurevych}.} \bibinfo{year}{2019}\natexlab{}.
\newblock \showarticletitle{Sentence-{BERT}: Sentence Embeddings using {S}iamese {BERT}-Networks}. In \bibinfo{booktitle}{\emph{Proceedings of the 2019 Conference on Empirical Methods in Natural Language Processing and the 9th International Joint Conference on Natural Language Processing (EMNLP-IJCNLP)}}, \bibfield{editor}{\bibinfo{person}{Kentaro Inui}, \bibinfo{person}{Jing Jiang}, \bibinfo{person}{Vincent Ng}, {and} \bibinfo{person}{Xiaojun Wan}} (Eds.). \bibinfo{publisher}{Association for Computational Linguistics}, \bibinfo{address}{Hong Kong, China}, \bibinfo{pages}{3982--3992}.
\newblock
\urldef\tempurl%
\url{https://doi.org/10.18653/v1/D19-1410}
\showDOI{\tempurl}


\bibitem[Ren et~al\mbox{.}(2023)]%
        {ren2023investigatingfactualknowledgeboundary}
\bibfield{author}{\bibinfo{person}{Ruiyang Ren}, \bibinfo{person}{Yuhao Wang}, \bibinfo{person}{Yingqi Qu}, \bibinfo{person}{Wayne~Xin Zhao}, \bibinfo{person}{Jing Liu}, \bibinfo{person}{Hao Tian}, \bibinfo{person}{Hua Wu}, \bibinfo{person}{Ji-Rong Wen}, {and} \bibinfo{person}{Haifeng Wang}.} \bibinfo{year}{2023}\natexlab{}.
\newblock \bibinfo{title}{Investigating the Factual Knowledge Boundary of Large Language Models with Retrieval Augmentation}.
\newblock
\newblock
\showeprint[arxiv]{2307.11019}~[cs.CL]


\bibitem[Sadasivan et~al\mbox{.}(2024)]%
        {sadasivan2024aigeneratedtextreliablydetected}
\bibfield{author}{\bibinfo{person}{Vinu~Sankar Sadasivan}, \bibinfo{person}{Aounon Kumar}, \bibinfo{person}{Sriram Balasubramanian}, \bibinfo{person}{Wenxiao Wang}, {and} \bibinfo{person}{Soheil Feizi}.} \bibinfo{year}{2024}\natexlab{}.
\newblock \bibinfo{title}{Can AI-Generated Text be Reliably Detected?}
\newblock
\newblock
\showeprint[arxiv]{2303.11156}~[cs.CL]


\bibitem[Salewski et~al\mbox{.}(2023)]%
        {salewski23-impersonation}
\bibfield{author}{\bibinfo{person}{Leonard Salewski}, \bibinfo{person}{Stephan Alaniz}, \bibinfo{person}{Isabel Rio-Torto}, \bibinfo{person}{Eric Schulz}, {and} \bibinfo{person}{Zeynep Akata}.} \bibinfo{year}{2023}\natexlab{}.
\newblock \showarticletitle{In-Context Impersonation Reveals Large Language Models\textquotesingle Strengths and Biases}. In \bibinfo{booktitle}{\emph{Advances in Neural Information Processing Systems}}, \bibfield{editor}{\bibinfo{person}{A.~Oh}, \bibinfo{person}{T.~Naumann}, \bibinfo{person}{A.~Globerson}, \bibinfo{person}{K.~Saenko}, \bibinfo{person}{M.~Hardt}, {and} \bibinfo{person}{S.~Levine}} (Eds.), Vol.~\bibinfo{volume}{36}. \bibinfo{publisher}{Curran Associates, Inc.}, \bibinfo{pages}{72044--72057}.
\newblock
\urldef\tempurl%
\url{https://proceedings.neurips.cc/paper_files/paper/2023/file/e3fe7b34ba4f378df39cb12a97193f41-Paper-Conference.pdf}
\showURL{%
\tempurl}


\bibitem[Spitale et~al\mbox{.}(2023)]%
        {spitale23-disinforms-better}
\bibfield{author}{\bibinfo{person}{Giovanni Spitale}, \bibinfo{person}{Nikola Biller-Andorno}, {and} \bibinfo{person}{Federico Germani}.} \bibinfo{year}{2023}\natexlab{}.
\newblock \showarticletitle{AI model GPT-3 (dis)informs us better than humans}.
\newblock \bibinfo{journal}{\emph{Science Advances}} \bibinfo{volume}{9}, \bibinfo{number}{26} (\bibinfo{year}{2023}), \bibinfo{pages}{eadh1850}.
\newblock
\urldef\tempurl%
\url{https://doi.org/10.1126/sciadv.adh1850}
\showDOI{\tempurl}


\bibitem[Stiff and Johansson(2022)]%
        {Stiff-DetectingComputer-generatedDisinformation-2022c}
\bibfield{author}{\bibinfo{person}{Harald Stiff} {and} \bibinfo{person}{Fredrik Johansson}.} \bibinfo{year}{2022}\natexlab{}.
\newblock \showarticletitle{Detecting computer-generated disinformation}.
\newblock \bibinfo{journal}{\emph{Int. J. Data Sci. Anal.}} \bibinfo{volume}{13}, \bibinfo{number}{4} (\bibinfo{date}{May} \bibinfo{year}{2022}), \bibinfo{pages}{363--383}.
\newblock
\showISSN{2364-415X,2364-4168}
\urldef\tempurl%
\url{https://doi.org/10.1007/s41060-021-00299-5}
\showDOI{\tempurl}


\bibitem[Tourille et~al\mbox{.}(2022)]%
        {Tourille-AutomaticDetectionTweets-2022q}
\bibfield{author}{\bibinfo{person}{Julien Tourille}, \bibinfo{person}{Babacar Sow}, {and} \bibinfo{person}{Adrian Popescu}.} \bibinfo{year}{2022}\natexlab{}.
\newblock \showarticletitle{Automatic detection of bot-generated tweets}. In \bibinfo{booktitle}{\emph{Proceedings of the 1st International Workshop on Multimedia AI against Disinformation}}. \bibinfo{publisher}{ACM}, \bibinfo{address}{New York, NY, USA}, \bibinfo{pages}{44–51}.
\newblock
\urldef\tempurl%
\url{https://doi.org/10.1145/3512732.3533584}
\showDOI{\tempurl}


\bibitem[Uchendu et~al\mbox{.}(2023)]%
        {uchendu23_attributionreview}
\bibfield{author}{\bibinfo{person}{Adaku Uchendu}, \bibinfo{person}{Thai Le}, {and} \bibinfo{person}{Dongwon Lee}.} \bibinfo{year}{2023}\natexlab{}.
\newblock \showarticletitle{Attribution and Obfuscation of Neural Text Authorship: A Data Mining Perspective}.
\newblock \bibinfo{journal}{\emph{SIGKDD Explor. Newsl.}} \bibinfo{volume}{25}, \bibinfo{number}{1} (\bibinfo{date}{jul} \bibinfo{year}{2023}), \bibinfo{pages}{1–18}.
\newblock
\showISSN{1931-0145}
\urldef\tempurl%
\url{https://doi.org/10.1145/3606274.3606276}
\showDOI{\tempurl}


\bibitem[Uchendu et~al\mbox{.}(2020)]%
        {uchendu-etal-2020-authorship}
\bibfield{author}{\bibinfo{person}{Adaku Uchendu}, \bibinfo{person}{Thai Le}, \bibinfo{person}{Kai Shu}, {and} \bibinfo{person}{Dongwon Lee}.} \bibinfo{year}{2020}\natexlab{}.
\newblock \showarticletitle{Authorship Attribution for Neural Text Generation}. In \bibinfo{booktitle}{\emph{Proceedings of the 2020 Conference on Empirical Methods in Natural Language Processing (EMNLP)}}, \bibfield{editor}{\bibinfo{person}{Bonnie Webber}, \bibinfo{person}{Trevor Cohn}, \bibinfo{person}{Yulan He}, {and} \bibinfo{person}{Yang Liu}} (Eds.). \bibinfo{publisher}{Association for Computational Linguistics}, \bibinfo{address}{Online}, \bibinfo{pages}{8384--8395}.
\newblock
\urldef\tempurl%
\url{https://doi.org/10.18653/v1/2020.emnlp-main.673}
\showDOI{\tempurl}


\bibitem[Vykopal et~al\mbox{.}(2024)]%
        {Vykopal-DisinformationCapabilitiesModels-2023c}
\bibfield{author}{\bibinfo{person}{Ivan Vykopal}, \bibinfo{person}{Matúš Pikuliak}, \bibinfo{person}{Ivan Srba}, \bibinfo{person}{Robert Moro}, \bibinfo{person}{Dominik Macko}, {and} \bibinfo{person}{Maria Bielikova}.} \bibinfo{year}{2024}\natexlab{}.
\newblock \bibinfo{title}{Disinformation Capabilities of Large Language Models}.
\newblock
\newblock
\showeprint[arxiv]{2311.08838}~[cs.CL]


\bibitem[Wang et~al\mbox{.}(2020)]%
        {wang-minilm-2020}
\bibfield{author}{\bibinfo{person}{Wenhui Wang}, \bibinfo{person}{Furu Wei}, \bibinfo{person}{Li Dong}, \bibinfo{person}{Hangbo Bao}, \bibinfo{person}{Nan Yang}, {and} \bibinfo{person}{Ming Zhou}.} \bibinfo{year}{2020}\natexlab{}.
\newblock \showarticletitle{MiniLM: Deep Self-Attention Distillation for Task-Agnostic Compression of Pre-Trained Transformers}. In \bibinfo{booktitle}{\emph{Proceedings of the 34th International Conference on Neural Information Processing Systems}} (Online), \bibfield{editor}{\bibinfo{person}{H.~Larochelle}, \bibinfo{person}{M.~Ranzato}, \bibinfo{person}{R.~Hadsell}, \bibinfo{person}{M.F. Balcan}, {and} \bibinfo{person}{H.~Lin}} (Eds.), Vol.~\bibinfo{volume}{33}. \bibinfo{publisher}{Curran Associates, Inc.}, \bibinfo{address}{Red Hook, NY, USA}, \bibinfo{pages}{5776--5788}.
\newblock
\urldef\tempurl%
\url{https://proceedings.neurips.cc/paper_files/paper/2020/file/3f5ee243547dee91fbd053c1c4a845aa-Paper.pdf}
\showURL{%
\tempurl}


\bibitem[Weber-Wulff et~al\mbox{.}(2023)]%
        {Weber-Wulff-TestingDetectionText-2023a}
\bibfield{author}{\bibinfo{person}{Debora Weber-Wulff}, \bibinfo{person}{Alla Anohina-Naumeca}, \bibinfo{person}{Sonja Bjelobaba}, \bibinfo{person}{Tomáš Foltýnek}, \bibinfo{person}{Jean Guerrero-Dib}, \bibinfo{person}{Olumide Popoola}, \bibinfo{person}{Petr Šigut}, {and} \bibinfo{person}{Lorna Waddington}.} \bibinfo{year}{2023}\natexlab{}.
\newblock \showarticletitle{Testing of detection tools for {AI}-generated text}.
\newblock \bibinfo{journal}{\emph{Int. J. Educ. Integr.}} \bibinfo{volume}{19}, \bibinfo{number}{1} (\bibinfo{date}{Dec.} \bibinfo{year}{2023}).
\newblock
\showISSN{1833-2595}
\urldef\tempurl%
\url{https://doi.org/10.1007/s40979-023-00146-z}
\showDOI{\tempurl}


\bibitem[Wei et~al\mbox{.}(2023)]%
        {wei23-jailbroken}
\bibfield{author}{\bibinfo{person}{Alexander Wei}, \bibinfo{person}{Nika Haghtalab}, {and} \bibinfo{person}{Jacob Steinhardt}.} \bibinfo{year}{2023}\natexlab{}.
\newblock \showarticletitle{Jailbroken: How Does LLM Safety Training Fail?}. In \bibinfo{booktitle}{\emph{Advances in Neural Information Processing Systems}} (New Orleans, LA, USA), \bibfield{editor}{\bibinfo{person}{A.~Oh}, \bibinfo{person}{T.~Naumann}, \bibinfo{person}{A.~Globerson}, \bibinfo{person}{K.~Saenko}, \bibinfo{person}{M.~Hardt}, {and} \bibinfo{person}{S.~Levine}} (Eds.), Vol.~\bibinfo{volume}{36}. \bibinfo{publisher}{Curran Associates, Inc.}, \bibinfo{address}{Red Hook, NY, USA}, \bibinfo{pages}{80079--80110}.
\newblock
\urldef\tempurl%
\url{https://proceedings.neurips.cc/paper_files/paper/2023/file/fd6613131889a4b656206c50a8bd7790-Paper-Conference.pdf}
\showURL{%
\tempurl}


\bibitem[Wei et~al\mbox{.}(2022)]%
        {NEURIPS2022_9d560961}
\bibfield{author}{\bibinfo{person}{Jason Wei}, \bibinfo{person}{Xuezhi Wang}, \bibinfo{person}{Dale Schuurmans}, \bibinfo{person}{Maarten Bosma}, \bibinfo{person}{Brian Ichter}, \bibinfo{person}{Fei Xia}, \bibinfo{person}{Ed~H. Chi}, \bibinfo{person}{Quoc~V. Le}, {and} \bibinfo{person}{Denny Zhou}.} \bibinfo{year}{2022}\natexlab{}.
\newblock \showarticletitle{Chain-of-Thought Prompting Elicits Reasoning in Large Language Models}. In \bibinfo{booktitle}{\emph{Proceedings of the 36th International Conference on Neural Information Processing Systems}} (New Orleans, LA, USA), \bibfield{editor}{\bibinfo{person}{S.~Koyejo}, \bibinfo{person}{S.~Mohamed}, \bibinfo{person}{A.~Agarwal}, \bibinfo{person}{D.~Belgrave}, \bibinfo{person}{K.~Cho}, {and} \bibinfo{person}{A.~Oh}} (Eds.), Vol.~\bibinfo{volume}{35}. \bibinfo{publisher}{Curran Associates, Inc.}, \bibinfo{address}{Red Hook, NY, USA}, \bibinfo{pages}{24824--24837}.
\newblock
\urldef\tempurl%
\url{https://proceedings.neurips.cc/paper_files/paper/2022/file/9d5609613524ecf4f15af0f7b31abca4-Paper-Conference.pdf}
\showURL{%
\tempurl}


\bibitem[Wilby et~al\mbox{.}(2023)]%
        {wilby-etal-2023-gate}
\bibfield{author}{\bibinfo{person}{David Wilby}, \bibinfo{person}{Twin Karmakharm}, \bibinfo{person}{Ian Roberts}, \bibinfo{person}{Xingyi Song}, {and} \bibinfo{person}{Kalina Bontcheva}.} \bibinfo{year}{2023}\natexlab{}.
\newblock \showarticletitle{{GATE} Teamware 2: An open-source tool for collaborative document classification annotation}. In \bibinfo{booktitle}{\emph{Proceedings of the 17th Conference of the European Chapter of the Association for Computational Linguistics: System Demonstrations}}, \bibfield{editor}{\bibinfo{person}{Danilo Croce} {and} \bibinfo{person}{Luca Soldaini}} (Eds.). \bibinfo{publisher}{Association for Computational Linguistics}, \bibinfo{address}{Dubrovnik, Croatia}, \bibinfo{pages}{145--151}.
\newblock
\urldef\tempurl%
\url{https://doi.org/10.18653/v1/2023.eacl-demo.17}
\showDOI{\tempurl}


\bibitem[Wu et~al\mbox{.}(2024)]%
        {wu2024surveyllmgeneratedtextdetection}
\bibfield{author}{\bibinfo{person}{Junchao Wu}, \bibinfo{person}{Shu Yang}, \bibinfo{person}{Runzhe Zhan}, \bibinfo{person}{Yulin Yuan}, \bibinfo{person}{Derek~F. Wong}, {and} \bibinfo{person}{Lidia~S. Chao}.} \bibinfo{year}{2024}\natexlab{}.
\newblock \bibinfo{title}{A Survey on LLM-Generated Text Detection: Necessity, Methods, and Future Directions}.
\newblock
\newblock
\showeprint[arxiv]{2310.14724}~[cs.CL]


\bibitem[Zellers et~al\mbox{.}(2019)]%
        {Zellers-DefendingAgainstNews-2019c}
\bibfield{author}{\bibinfo{person}{Rowan Zellers}, \bibinfo{person}{Ari Holtzman}, \bibinfo{person}{Hannah Rashkin}, \bibinfo{person}{Yonatan Bisk}, \bibinfo{person}{Ali Farhadi}, \bibinfo{person}{Franziska Roesner}, {and} \bibinfo{person}{Yejin Choi}.} \bibinfo{year}{2019}\natexlab{}.
\newblock \showarticletitle{Defending Against Neural Fake News}. In \bibinfo{booktitle}{\emph{Proceedings of the 33rd International Conference on Neural Information Processing Systems}} (Vancouver, Canada), \bibfield{editor}{\bibinfo{person}{H~Wallach}, \bibinfo{person}{H~Larochelle}, \bibinfo{person}{A~Beygelzimer}, \bibinfo{person}{F~d\textquotesingle Alché-Buc}, \bibinfo{person}{E~Fox}, {and} \bibinfo{person}{R~Garnett}} (Eds.), Vol.~\bibinfo{volume}{32}. \bibinfo{publisher}{Curran Associates, Inc.}, \bibinfo{address}{Red Hook, NY, USA}, \bibinfo{pages}{9054--9065}.
\newblock
\urldef\tempurl%
\url{https://proceedings.neurips.cc/paper_files/paper/2019/file/3e9f0fc9b2f89e043bc6233994dfcf76-Paper.pdf}
\showURL{%
\tempurl}


\bibitem[Zhang et~al\mbox{.}(2023)]%
        {zhang2023sirenssongaiocean}
\bibfield{author}{\bibinfo{person}{Yue Zhang}, \bibinfo{person}{Yafu Li}, \bibinfo{person}{Leyang Cui}, \bibinfo{person}{Deng Cai}, \bibinfo{person}{Lemao Liu}, \bibinfo{person}{Tingchen Fu}, \bibinfo{person}{Xinting Huang}, \bibinfo{person}{Enbo Zhao}, \bibinfo{person}{Yu Zhang}, \bibinfo{person}{Yulong Chen}, \bibinfo{person}{Longyue Wang}, \bibinfo{person}{Anh~Tuan Luu}, \bibinfo{person}{Wei Bi}, \bibinfo{person}{Freda Shi}, {and} \bibinfo{person}{Shuming Shi}.} \bibinfo{year}{2023}\natexlab{}.
\newblock \bibinfo{title}{Siren's Song in the AI Ocean: A Survey on Hallucination in Large Language Models}.
\newblock
\newblock
\showeprint[arxiv]{2309.01219}~[cs.CL]


\bibitem[Zheng et~al\mbox{.}(2023)]%
        {zheng2023doeschatgptfallshort}
\bibfield{author}{\bibinfo{person}{Shen Zheng}, \bibinfo{person}{Jie Huang}, {and} \bibinfo{person}{Kevin Chen-Chuan Chang}.} \bibinfo{year}{2023}\natexlab{}.
\newblock \bibinfo{title}{Why Does ChatGPT Fall Short in Providing Truthful Answers?}
\newblock
\newblock
\showeprint[arxiv]{2304.10513}~[cs.CL]


\end{thebibliography}
\end{document}